\documentclass{article}

\PassOptionsToPackage{numbers,sort&compress}{natbib}

\usepackage[preprint]{neurips_2026}

\usepackage[T1]{fontenc}
\usepackage{hyperref}
\usepackage{url}
\usepackage{amsfonts}
\usepackage{amssymb}
\usepackage{amsmath}
\usepackage{nicefrac}
\usepackage{microtype}
\usepackage{array}

\usepackage{xspace}
\usepackage{mathtools} 
\usepackage{bm}
\usepackage{enumitem}
\usepackage{textcomp}
\usepackage{xcolor}
\usepackage{changepage}
\usepackage{diagbox}

\usepackage{graphicx}
\usepackage{subcaption}
\usepackage{tikz}
\usepackage{caption}

\usetikzlibrary{arrows.meta,positioning,fit,calc,shapes.geometric,shapes.symbols,backgrounds}

\usepackage{tabularx}
\newcolumntype{Y}{>{\centering\arraybackslash}X}
\usepackage{multirow}
\usepackage{booktabs}
\usepackage{makecell}
\usepackage[figuresleft]{rotating}
\usepackage[flushleft]{threeparttable}
\usepackage{xcolor}
\usepackage{colortbl}
\usepackage{algorithmicx}
\usepackage{algorithm}
\usepackage[noend]{algpseudocode}

\algnewcommand\algorithmicinput{\textbf{Input:}}
\algnewcommand\Input{\item[\algorithmicinput]}
\algnewcommand\algorithmicoutput{\textbf{Output:}}
\algnewcommand\Output{\item[\algorithmicoutput]}

\usepackage{tcolorbox} 
\tcbuselibrary{listings,breakable,skins}

\usepackage[utf8]{inputenc} 
\usepackage{tcolorbox} 
\usepackage{soul} 
\usepackage{pifont} 
\usepackage{bbding} 
\usepackage[textsize=footnotesize, linecolor=magenta, bordercolor=magenta, backgroundcolor=white, textwidth=35pt]{todonotes}
\usepackage{marginnote}

\usepackage[noindentafter]{titlesec}
\titlespacing\section{0pt}{3pt}{3pt} 
\titlespacing\subsection{0pt}{3pt}{3pt}
\titlespacing\subsubsection{0pt}{5pt}{5pt}

\title{Hide to Guide: Learning via Semantic Masking}

\definecolor{mitblue}{rgb}{0.88,0.95,0.96}

\newcommand{\experttrace}{\tau}
\newcommand{\maskedexperttrace}{\widetilde{\tau}}
\newcommand{\method}{SMEPO}

\author{%
\parbox{\linewidth}{\centering
{\bfseries
\makebox[\textwidth][c]{%
\begin{tabular}{@{}c@{\hspace{1.25em}}
                c@{\hspace{1.25em}}
                c@{\hspace{1.25em}}
                c@{\hspace{1.25em}}
                c@{}}
Ruitao Liu\textsuperscript{1}\thanks{Equal contribution.} &
Qinghao Hu\textsuperscript{1}\footnotemark[1] &
Alex Hu\textsuperscript{1} &
Yecheng Wu\textsuperscript{2} &
Shang Yang\textsuperscript{1}
\end{tabular}}%
\\[0.6ex]
\makebox[\textwidth][c]{%
\begin{tabular}{@{}c@{\hspace{1.25em}}
                c@{\hspace{1.25em}}
                c@{\hspace{1.25em}}
                c@{}}
Luke J. Huang\textsuperscript{1} &
Zhuoyang Zhang\textsuperscript{1} &
Han Cai\textsuperscript{2} &
Song Han\textsuperscript{1,2}
\end{tabular}}%
}
\\[1.0ex]
{\normalfont
\textsuperscript{1}MIT \quad
\textsuperscript{2}NVIDIA
}
}}

\begin{document}

\maketitle

\begin{abstract}

Reinforcement learning with verifiable rewards (RLVR) has become a powerful paradigm for improving language models on reasoning-intensive tasks, but its effectiveness is often limited by exploration. For example, models often fail on hard problems, leaving little useful reward signal. External expert traces offer a natural source of guidance, yet they may also expose reward-relevant content along the critical path to the verifier target, such as final answers, intermediate values, executable implementations, or answer-related entities. This content can create an unintended reward hacking channel, allowing the policy to obtain reward by copying the trace rather than learning the underlying reasoning or agentic behavior. 
Existing guided-RL methods reduce this risk by using partial trajectories, but they mainly control how much expert information is shown heuristically rather than which parts should be hidden. To this end, we propose Semantic Masked Expert Policy Optimization (SMEPO), a fine-grained semantic masking strategy for expert-guided RLVR. Instead of truncating traces coarsely or revealing them unchanged, SMEPO masks reward-relevant semantic spans along the critical path while preserving the expert’s decomposition, plan, and procedural structure. This turns hard problems from reasoning from scratch into a fill-in-the-blank process: the policy can follow the expert’s problem-solving route, but must still reconstruct the missing values, code, or entities by itself. SMEPO is simple to apply and requires no changes to the reward function or RL objective.
Across diverse domains, including math, code, and agentic search, SMEPO improves accuracy by up to 3.2 points over GRPO and reduces training time by up to 4.2$\times$. 
The code is available at \url{https://github.com/mit-han-lab/SMEPO}.
\end{abstract}

\begin{figure*}[t]
    \centering
    \includegraphics[width=\textwidth]{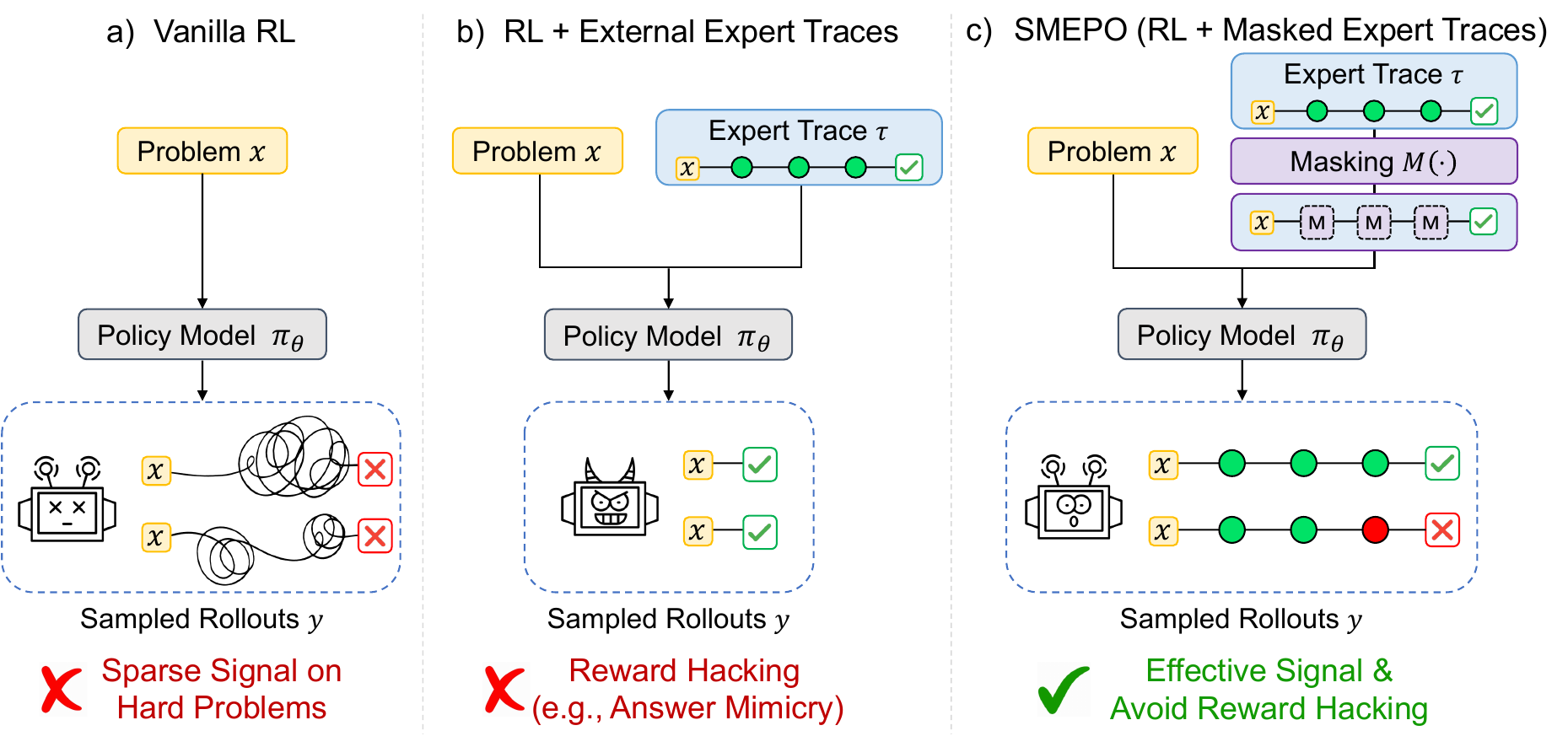}
\caption{\textbf{Overview of SMEPO.}
On hard reasoning problems, a) vanilla RL often receives sparse rewards because the initial policy fails to discover useful solution trajectories. b) Full expert traces can provide useful trajectories, but may also reveal verifier-aligned content along the solution path, such as intermediate values, final answers, code blocks, or answer-related entities. RL conditioned on the full trace may therefore exploit these exposed cues rather than learn the underlying reasoning process. c) SMEPO masks reward-relevant spans along the critical path while leaving the surrounding trace intact. The resulting fill-in-the-blank trace helps exploration while hiding verifier-aligned content.}
    \label{fig:teaser}
\end{figure*}

\section{Introduction}

Reinforcement learning with verifiable rewards (RLVR) has emerged as a prominent paradigm for enhancing large language models in complex domains such as mathematics, coding, and agentic search \citep{deepseekmath,deepseekr1,reinforcepp,dapo,oneshotrlvr,understandr1zero,deepresearcher}. Unlike methods that rely on imitating static datasets, RLVR directly optimizes model behavior using automated verification. However, relying purely on self-exploration presents a critical bottleneck. On hard tasks, early rollouts frequently fail, resulting in sparse rewards that provide insufficient learning signals for the policy \citep{sparse}. Consequently, practical RLVR systems typically require auxiliary guidance to steer the exploration process toward viable solution trajectories.

To provide this necessary guidance, a natural approach is to condition the RL process on expert traces, such as chain-of-thought rationales, solution trajectories, or hints. This concept has inspired a growing family of guided RL methods designed to enhance exploration \citep{ghpo,trapo,pope,luffy,guide}. While these approaches share the core intuition that expert signals can help policies navigate spaces they cannot explore autonomously, they notably avoid exposing the model to complete, unmodified expert traces. Instead, they rely on fragmented or altered signals, such as adaptive hints, partial prefixes, and rewritten trajectories. We argue that this deliberate design choice exposes a fundamental failure mode in standard expert-trace conditioning.

We attribute this failure to the dual nature of expert traces. They contain useful procedural cues, such as equation templates, solution plans, and query decomposition, but also expose reward-relevant content along the critical path to the verifier target. In math, this includes final answers and intermediate numerical values; in code, complete executable programs; and in agentic search, answer-relevant entities and query pivots. Since RLVR typically evaluates only the final outcome through domain-specific verifiers \citep{deepseekmath,deepseekr1,reft,dapo}, the policy may obtain reward by relying on these exposed cues rather than learning the reasoning, coding, or search process that produced them. In this way, full expert traces can turn guidance into an unintended reward hacking channel.

This perspective suggests a different design principle for guided RL: preserve the procedural scaffolding of expert traces while hiding verifier-aligned content along the critical path. Partial trajectories and adaptive hints can help because they reduce access to reward-relevant content, but when high-quality expert data is scarce, discarding large parts of the trace can also waste valuable guidance. In particular, it can remove or truncate the process-level information that makes expert traces useful, such as the step-by-step plan, intermediate subgoals, and problem-solving patterns.

We propose Semantic Masked Expert Policy Optimization (SMEPO), a semantically masked expert-guidance method that converts expert traces into fill-in-the-blank-style guidance. Instead of appending the full trace unchanged, SMEPO masks semantic spans that directly reveal the verifier-checked outcome or lie on the critical path to obtaining it, while preserving procedural guidance. We instantiate this principle in three representative RLVR domains, while the masking targets are not limited to these domains. For math, we mask numeric content in the expert trace; for code, we mask expert solution code; and for agentic search, we mask answer-relevant entities in expert search traces. The resulting trace preserves the expert's solution process, including decomposition, intermediate subgoals, and reasoning flow, while removing verifier-aligned content along the critical path. SMEPO introduces a guidance-level intervention: it reshapes the expert trace seen by the policy while keeping the verifier, reward function, and RL objective unchanged. This makes SMEPO easy to integrate into GRPO-style RLVR pipelines and combine with other guidance strategies.

Experiments show that SMEPO improves over both vanilla GRPO and direct expert conditioning. In our main 1k-example-per-domain setting, SMEPO improves Qwen3-8B-Base from 47.6 to 50.3 on math, 56.5 to 59.0 on code, and 45.6 to 48.8 on agentic search. The gains transfer to additional math models, improving Qwen2.5-7B from 42.4 to 44.5 and DeepSeek-R1-Distill-Qwen-7B from 61.8 to 62.6, and SMEPO reaches hard-math targets earlier than vanilla GRPO. In a GHPO-based setting using the original GHPO data \citep{ghpo}, adding SMEPO further improves the math average from 44.0 to 45.6. Diagnostic ablations support our explanation: random masking fails, fixed-prefix exposure is unstable, and progressively more targeted removal of reward-relevant content works best, indicating that the harmful signal is structured and reward-relevant rather than merely longer or noisier guidance.

We summarize our contributions as follows:
\begin{itemize}[leftmargin=2em]
    \item We identify a critical failure mode in expert-guided RLVR. We characterize how policies bypass learning the underlying reasoning by exploiting verifier-aligned content (e.g., numerical values, entities) exposed in full expert traces across math, code, and agentic search domains.
    \item We propose SMEPO, a lightweight, guidance-level intervention. By masking reward-relevant spans while preserving the procedural scaffolding, SMEPO converts expert traces into effective fill-in-the-blank guidance without leaking solutions.
    \item We demonstrate that SMEPO significantly improves over GRPO and direct expert conditioning across multiple LLMs and diverse domains, while accelerating convergence on hard tasks.
\end{itemize}

\section{Method}
\label{sec:method}

SMEPO is motivated by a tension in expert-guided RLVR: expert traces aid exploration but may also reveal content on the critical path to verifier success. Appendix~\ref{app:case-studies} illustrates the two motivating failure modes: vanilla GRPO struggles to discover useful trajectories, whereas full-expert-trace conditioning exposes content that the policy can exploit during RL. We address this by separating reusable procedural guidance from reward-relevant content through lightweight semantic masking.

We consider reinforcement learning with verifiable rewards (RLVR). Given a problem prompt $x$, the policy $\pi_\theta$ generates a response $y$, and a domain-specific verifier assigns a reward $R(x,y)$. Unlike supervised imitation, the policy is optimized only through verifier feedback.

In expert-trace-guided RL, each training problem is additionally paired with an externally generated trace $\experttrace$ from a stronger model, such as a reasoning trace for math, a reference solution for code, or a search trajectory for agentic QA. A direct use of this trace is full-expert-trace conditioning:
\begin{equation}
    y \sim \pi_\theta(\cdot \mid x, \experttrace),
\end{equation}
while optimizing the same verifier reward $R(x,y)$. The trace is visible to the policy during rollout generation, but is not part of the generated answer; the verifier still scores only $y$. Thus, any benefit or harm from $\experttrace$ comes through how it changes rollout behavior.

This raises the central design question of SMEPO: how can we preserve the exploration benefits of expert traces while hiding reward-relevant content along the critical path? We formalize this tension by viewing an expert trace as containing two types of information:
\begin{equation}
    \experttrace = \big(\experttrace^{\mathrm{g}}, \experttrace^{\mathrm{r}}\big).
\end{equation}
Here $\experttrace^{\mathrm{g}}$ denotes reusable guidance, such as solution organization, decomposition patterns, equation templates, control-flow logic, or step-by-step progression. In contrast, $\experttrace^{\mathrm{r}}$ denotes content whose surface form is closely tied to verifier success, such as final answers, key intermediate values, executable implementations, or critical entities along a search path. When exposed unchanged, this reward-relevant content can become an unintended reward channel: the policy may rely on the trace to reach high verifier reward without learning the underlying reasoning, coding, or search process.

Motivated by this decomposition, SMEPO constructs a masked trace
\begin{equation}
    \maskedexperttrace = M(\experttrace),
\end{equation}
where $M$ preserves the order, formatting, and procedural organization of the trace while replacing reward-relevant spans with mask placeholders. The policy is then trained as
\begin{equation}
    y \sim \pi_\theta(\cdot \mid x,\maskedexperttrace),
\end{equation}
using the same verifier reward $R(x,y)$ and the same GRPO pipeline as the baseline. Abstractly, this corresponds to optimizing
\begin{equation}
    \max_{\theta}\;
    \mathbb{E}_{(x,\experttrace)\sim \mathcal{D},\; y\sim \pi_{\theta}(\cdot \mid x,\maskedexperttrace)}
    \left[R(x,y)\right].
\end{equation}

\begin{figure*}[t]
    \centering
    \includegraphics[width=\textwidth]{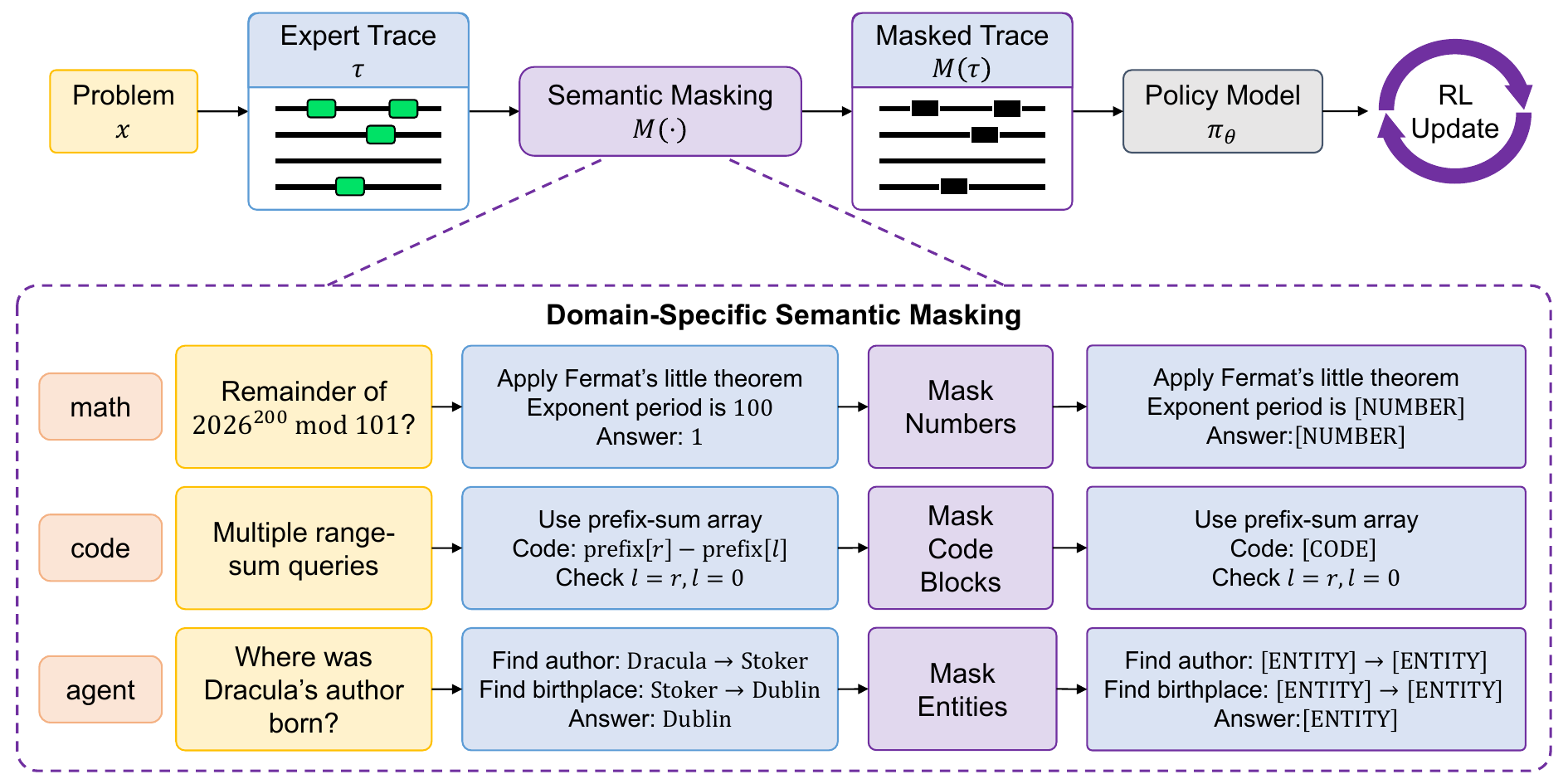}
    \caption{\textbf{Training pipeline for SMEPO.} We detect reward-relevant spans in the expert trace with domain-specific masking rules, replace those spans with mask placeholders, and then run the same GRPO loop conditioned on the original problem and the masked trace.}
    \label{fig:pipeline}
\end{figure*}

Figure~\ref{fig:pipeline} summarizes the workflow. SMEPO applies masking before policy optimization, so the learner receives a modified guidance trace throughout rollout generation. Although the concrete masking rules differ by domain, they follow the same principle: preserve expert guidance while suppressing content that either directly reveals the verifier target or exposes key intermediate information on the critical path to obtaining it.

\paragraph{Semantic masking.}
SMEPO masks reward-relevant content along the critical path to the verifier target while preserving the surrounding guidance. In RLVR, verifiers are domain-specific: math verifiers compare extracted numeric answers, code verifiers execute generated programs against tests, and agentic-search verifiers compare answer entities. SMEPO mirrors this verifier semantics at the guidance level: it masks semantic types that are directly checked by the verifier or reveal the path to what the verifier checks. Such spans can often be identified with lightweight domain-specific rules or semantic tools, such as number matching, code-block detection, named-entity recognition with spaCy~\citep{spacy}, or lightweight general-purpose NER models such as GLiNER~\citep{gliner}. We instantiate this principle in three representative domains below, with additional implementation details in Appendix~\ref{app:masking-details}.

\paragraph{Domain 1: math reasoning.}
For math, reward-relevant content often includes not only the final answer, but also repeated answer values and intermediate numeric values along the solution path. These numbers can reveal the target through copying, back-solving, or value-specific arithmetic. We therefore mask numeric spans in the expert trace, such as integers, decimals, and fractions. This preserves guidance such as decomposition steps, equation templates, and explanatory text, while requiring the policy to reconstruct the value-relevant parts of the solution itself.

\paragraph{Domain 2: code generation.}
For code, the most directly reward-relevant content is the expert program. A complete reference implementation can be copied, lightly adapted, or used as an executable template. We therefore mask expert code blocks while preserving surrounding non-code text. This removes exposed executable content while retaining useful guidance such as task interpretation, algorithmic plans, data-structure choices, corner-case discussions, and high-level debugging hints.

\paragraph{Domain 3: agentic search.}
For agentic search, expert trajectories can expose entities that are tightly coupled to verifier success. Expert trajectories may reveal intermediate entities, query pivots, supporting-page titles, and the final answer entity. Exposing these entities can leak the multi-hop route. We therefore apply entity-level masking over the expert search trace. For each trace, we identify salient entity spans from answer spans, emphasized terms, and named entities detected by spaCy~\citep{spacy}, and mask all occurrences within the same trace. This preserves the multi-step search trajectory while hiding entities that would otherwise reveal the search path or final answer.

\paragraph{Beyond these domains.} This principle naturally extends beyond our evaluated domains. The three domains studied here cover common RLVR regimes: long-form reasoning with closed-form answer verification, execution-based generation, and multi-round tool interaction. Many other RLVR tasks can be viewed as variants or combinations of these regimes, with different semantic content lying on the critical path to the answer or task completion. Applying SMEPO to a new domain therefore amounts to identifying this domain-specific content and masking it while preserving the surrounding procedural guidance. For example, interactive agents extend multi-round tool interaction to web navigation, mobile-use, and embodied-control tasks, where SMEPO could mask URLs, page titles, object names, UI labels, coordinates, or action arguments while preserving the high-level action plan. Thus, what changes across settings is the domain-specific semantic type along the critical path, not the masked-guidance principle itself.


In the three domains studied here, SMEPO masking is applied offline with lightweight preprocessing. Across \textit{all training data} for the three domains combined, preprocessing takes only \textit{37 seconds in total} on the CPU node. This one-time cost is negligible compared with rollout generation and model optimization; detailed per-domain measurements are reported in Appendix~\ref{app:method-overhead}.

Overall, SMEPO differs from coarse trace truncation by preserving the trace whenever possible and masking only spans selected by reward-relevant rules. Rather than discarding large portions of the expert trace, SMEPO keeps the surrounding procedural guidance intact while removing verifier-aligned content along the critical path. We test this design choice in Section~\ref{sec:ablation} by comparing against fixed-prefix exposure and same-budget random masking baselines.
\section{Experiments}
\label{sec:experiments}

\paragraph{Models.}
We evaluate three open-weight models: Qwen3-8B-Base~\citep{qwen3technicalreport}, Qwen2.5-7B~\citep{qwen25technicalreport}, and DeepSeek-R1-Distill-Qwen-7B~\citep{deepseekr1,qwen25technicalreport}. Qwen3-8B-Base is our main model for math, code, and agentic search. Qwen2.5-7B and DeepSeek-R1-Distill-Qwen-7B are used as additional math models to test whether the degradation from full expert traces and the gains from SMEPO persist across different model families and base reasoning capabilities. 

\paragraph{Evaluation tasks.}
We evaluate across math, code, and agentic search. For math, we use GSM8K~\citep{gsm8k}, MATH500~\citep{math500}, AIME25~\citep{aime25}, AIME26~\citep{aime26}, AMC~\citep{amc23}, and OlympiadBench~\citep{olympiadbench}, and report the unweighted average over the six benchmarks. For AIME25 and AIME26, we average accuracy over 32 sampled responses per problem to reduce variance. For code, we evaluate pass@1 on HumanEval~\citep{humaneval}, HumanEval+~\citep{humanevalplus}, and LiveCodeBench~\citep{lcb}. We also include a commonly used 131-problem LiveCodeBench v6 subset evaluation in Appendix~\ref{app:lcb-v6}, where the same qualitative trend holds. For agentic search, we evaluate accuracy on Bamboogle~\citep{bamboogle}. We also report 5-shot MMLU-Overall~\citep{mmlu_1,mmlu_2} in Appendix~\ref{app:mmlu} as a general-domain sanity check.

\paragraph{Training data and expert traces.}
The main math and code experiments use separate 1k-example subsets from OpenThoughts-114k~\citep{openthoughts}. Both subsets contain DeepSeek-R1 expert traces. In the main math experiments with Qwen3-8B-Base, we additionally evaluate Gemini-3-Pro-Preview traces to test robustness to the expert source. For agentic search, we use a 1k-example subset of MiroVerse HotpotQA search trajectories~\citep{miroverse}. Within each experimental block, all methods use the same training problems, verifier, rollout budget, and expert source. Appendix~\ref{app:large-scale-89k} reports an additional large-scale math experiment using 89k math examples from OpenThoughts.

\paragraph{Compared methods.}
We compare vanilla \textbf{GRPO} without expert traces; \textbf{SFT} on expert responses; and \textbf{SFT + RL}, which applies GRPO after SFT. For expert-trace guidance, \textbf{Expert} appends the full trace during GRPO, while \textbf{SMEPO} uses the same trace with reward-relevant spans masked. We also include \textbf{LUFFY}~\citep{luffy} and \textbf{GHPO}~\citep{ghpo}, both using the same expert traces as Expert and SMEPO.

\begin{table*}[!t]
\small
\centering
\renewcommand{\arraystretch}{1.08}
\setlength{\tabcolsep}{5pt}

\begin{tabularx}{\textwidth}{lYYYYYYY}
\toprule
\rowcolor{blue!22}
\multicolumn{8}{c}{\textbf{Math}} \\
\midrule
Method & GSM8K & MATH500 & AIME25 & AIME26 & AMC & Olympiad & Avg. \\
\midrule
\rowcolor{blue!12}
\multicolumn{8}{c}{\textbf{Qwen3-8B-Base}} \\
\midrule
Qwen3-8B-Base              & 68.0 & 67.0 & 9.9 & 7.6 & 50.0 & 25.8 & 38.1 \\
GRPO                       & 91.3 & 75.2 & 12.5 & 12.3 & 65.0 & 29.5 & 47.6 \\
\midrule
SFT (DS)                   & 88.0 & 71.2 & 8.0 & 5.6 & 50.0 & 25.4 & 41.4 \\
SFT (DS) + RL              & 91.7 & \underline{76.8} & 15.0 & 10.4 & 62.5 & 29.7 & 47.7 \\
Expert (DS)                & 76.4 & 69.2 & 11.4 & 8.6 & 52.5 & 27.7 & 41.0 \\
LUFFY (DS)                 & 90.4 & 74.6 & 14.4 & 12.6 & 60.0 & 29.2 & 46.9 \\
GHPO (DS)                  & 79.5 & 73.6 & 13.5 & 10.0 & 60.0 & 29.2 & 44.3 \\
\rowcolor{mitblue}
SMEPO (DS)                 & 92.2 & 76.6 & \textbf{17.6} & 13.4 & \textbf{70.0} & \underline{32.0} & \textbf{50.3} \\
\midrule
SFT (Gemini)               & 91.7 & 71.2 & 10.2 & 9.0 & 52.5 & 25.2 & 43.3 \\
SFT (Gemini) + RL          & 91.6 & 73.0 & 13.2 & 10.6 & 60.0 & 28.0 & 46.1 \\
Expert (Gemini)            & 78.0 & 68.8 & 11.6 & 9.6 & 50.0 & 26.6 & 40.8 \\
LUFFY (Gemini)             & \textbf{92.4} & 74.6 & 15.2 & \textbf{14.7} & 65.0 & 30.7 & \underline{48.8} \\
GHPO (Gemini)              & 79.4 & 73.2 & 13.9 & 9.5 & 65.0 & 29.5 & 45.1 \\
\rowcolor{mitblue}
SMEPO (Gemini)             & \underline{92.3} & \textbf{78.6} & \underline{16.1} & \underline{14.4} & \underline{67.5} & \textbf{32.8} & \textbf{50.3} \\
\midrule
\rowcolor{blue!12}
\multicolumn{8}{c}{\textbf{Qwen2.5-7B}} \\
\midrule
Qwen2.5-7B                 & 46.2 & 55.8 & 4.2 & 5.4 & 35.0 & 20.0 & 27.8 \\
GRPO                       & \underline{87.6} & \underline{70.2} & \underline{9.4} & \underline{7.7} & \underline{55.0} & \underline{24.3} & \underline{42.4} \\
SFT (DS)                   & 87.0 & 66.8 & 7.3 & 5.0 & 42.5 & 22.1 & 38.5 \\
Expert (DS)                & 85.3 & 63.8 & 5.6 & 6.4 & 47.5 & 22.8 & 38.6 \\
\rowcolor{mitblue}
SMEPO (DS)                  & \textbf{89.0} & \textbf{72.4} & \textbf{10.8} & \textbf{9.7} & \textbf{60.0} & \textbf{25.1} & \textbf{44.5} \\

\midrule
\rowcolor{blue!12}
\multicolumn{8}{c}{\textbf{DeepSeek-R1-Distill-Qwen-7B}} \\
\midrule
R1-Distill-Qwen-7B         & 83.5 & 81.0 & 32.6 & 33.3 & 82.5 & 35.3 & 58.0 \\
GRPO                       & 86.4 & \underline{84.2} & 34.0 & \underline{38.8} & \textbf{90.0} & \underline{37.2} & \underline{61.8} \\
Expert (DS)                & \underline{87.3} & 81.0 & \underline{34.1} & 38.4 & \underline{85.0} & 36.8 & 60.4 \\
\rowcolor{mitblue}
SMEPO (DS)                  & \textbf{87.5} & \textbf{85.4} & \textbf{34.7} & \textbf{39.7} & \textbf{90.0} & \textbf{38.3} & \textbf{62.6} \\
\bottomrule
\end{tabularx}

\vspace{0.8em}

\noindent\makebox[\textwidth][c]{%
\begin{minipage}[c]{0.65\textwidth}
\centering
\begin{tabularx}{\textwidth}{lYYYY}
\toprule
\rowcolor{blue!22}
\multicolumn{5}{c}{\textbf{Code: Qwen3-8B-Base}} \\
\midrule
Method & HumanEval & HumanEval+ & LCB & Avg. \\
\midrule
Qwen3-8B-Base & 63.4 & 55.5 & 35.0 & 51.3 \\
GRPO          & \underline{67.7} & 59.8 & \underline{42.0} & \underline{56.5} \\
SFT           & 67.1 & \underline{60.4} & 28.8 & 52.1 \\
Expert        & 66.5 & 58.5 & \underline{42.0} & 55.7 \\
\rowcolor{mitblue}
SMEPO          & \textbf{69.5} & \textbf{62.2} & \textbf{45.3} & \textbf{59.0} \\
\bottomrule
\end{tabularx}
\end{minipage}%
\hfill
\begin{minipage}[c]{0.30\textwidth}
\centering
\begin{tabularx}{\textwidth}{lY}
\toprule
\rowcolor{blue!22}
\multicolumn{2}{c}{\textbf{Agent: Qwen3-8B-Base}} \\
\midrule
Method & Bamboogle \\
\midrule
Qwen3-8B-Base & 20.8 \\
GRPO          & \underline{45.6} \\
SFT           & 44.0 \\
Expert        & 23.2 \\
\rowcolor{mitblue}
SMEPO          & \textbf{48.8} \\
\bottomrule
\end{tabularx}
\end{minipage}%
}

\caption{\textbf{Main results}. (\textbf{top}) Math performance on Qwen3-8B-Base, Qwen2.5-7B, and DeepSeek-R1-Distill-Qwen-7B. The suffixes (DS) and (Gemini) indicate that the expert traces are sourced from DeepSeek-R1 and Gemini-3-Pro, respectively. (\textbf{bottom}) Code and agentic search performance on Qwen3-8B-Base. Across settings, naive full expert traces hurt or fail to help, while SMEPO consistently improves over both vanilla GRPO and naive expert-trace guidance. Best results are in bold; second-best results are underlined.}
\label{tab:main-results}
\end{table*}


\paragraph{Training and evaluation details.}
Unless otherwise noted, RL training uses 8 rollouts per prompt and runs for 400 optimization steps on 8$\times$H100 GPUs. We use a maximum response length of 12k for math and code. For search, which is multi-round, we use a maximum response length of 2048 tokens per round.
The learning rate is $1 \times 10^{-6}$ for math and code, and $8 \times 10^{-7}$ for search. Expert traces or masked traces are appended after the original problem as additional context and are used only during rollout generation. Rewards are computed from the model-generated answer using the same verifier as vanilla GRPO. Evaluation uses lm-evaluation-harness~\citep{eval-harness} for supported benchmarks; for benchmarks not supported by lm-evaluation-harness, evaluation details are included in Appendix~\ref{app:training-details}.

\subsection{Main results}

Table~\ref{tab:main-results} shows that complete expert traces consistently fail to improve vanilla GRPO, while masked expert traces improve performance across domains. In the main math setting with Qwen3-8B-Base, vanilla GRPO achieves a six-benchmark average of 47.6. Appending complete expert traces reduces the average to 41.0 with DeepSeek-R1 traces and 40.8 with Gemini traces. In contrast, SMEPO reaches 50.3 with both expert sources, improving over vanilla GRPO by 2.7 points and over full-expert-trace conditioning by 9.3--9.5 points.

The same trend holds across other models. On Qwen2.5-7B, full-expert-trace conditioning reduces the average from 42.4 to 38.6, while SMEPO improves it to 44.5. On DeepSeek-R1-Distill-Qwen-7B, full-expert-trace conditioning reduces the average from 61.8 to 60.4, while SMEPO improves it to 62.6. The smaller degradation on DeepSeek-R1-Distill-Qwen-7B likely reflects its stronger prior reasoning ability, which makes it less dependent on exposed expert content than base models.

\begin{figure*}[t]
    \centering
    \begin{subfigure}[t]{0.42\textwidth}
        \centering
        \includegraphics[width=\linewidth]{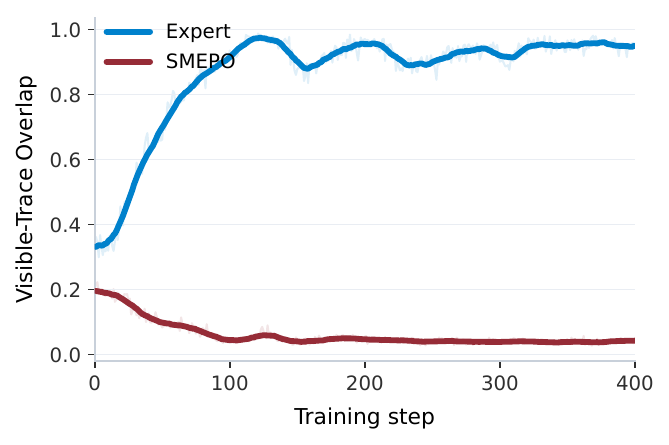}
        \caption{Visible-trace overlap.}
        \label{fig:visible-trace-overlap-curve}
    \end{subfigure}
    \hfill
    \begin{subfigure}[t]{0.42\textwidth}
        \centering
        \includegraphics[width=\linewidth]{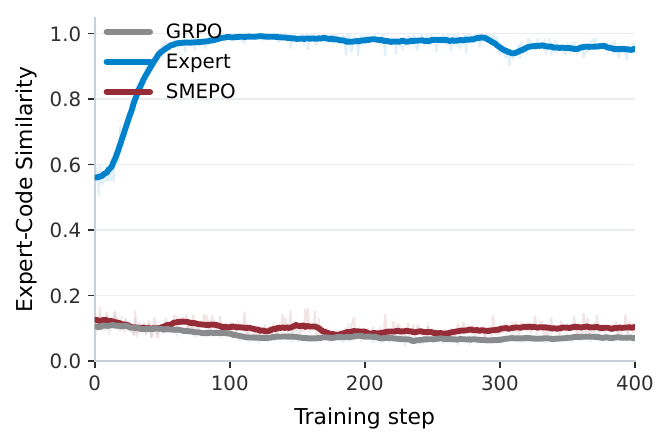}
        \caption{Expert-code similarity.}
        \label{fig:expert-code-similarity}
    \end{subfigure}
    \caption{\textbf{Code-copy diagnostics.}
    \textbf{Left:} \emph{Visible-trace overlap} captures copying from the expert content actually visible in the prompt.
    \textbf{Right:} \emph{Expert-code similarity} captures reuse of the original expert program. The left figure does not include GRPO since it does not involve expert traces.}
    \label{fig:code-copy-diagnostics}
\end{figure*}

For code, complete expert traces are less harmful than in math and agentic search, likely because exploiting them requires reproducing or adapting a longer executable solution rather than copying short reward-relevant content such as numbers or entities. On Qwen3-8B-Base, GRPO obtains a three-benchmark average of 56.5, full-expert-trace conditioning obtains 55.7, and SMEPO improves the average to 59.0. SMEPO also gives the best score on each reported code benchmark. Although the accuracy drop from full traces is smaller in code than in math or search, Figure~\ref{fig:code-copy-diagnostics} shows that full-expert-trace conditioning substantially increases copying from the visible expert solution and reuse of the original expert program, whereas SMEPO suppresses both effects. This indicates that complete code traces still bias the policy toward trace reuse, while SMEPO reduces this dependency by masking the executable solution span.

For agentic search, the degradation from complete expert traces is the largest. On Bamboogle, GRPO reaches 45.6, full-expert-trace conditioning drops to 23.2, and SMEPO improves to 48.8. This sharp drop suggests that full traces can suppress the policy's active use of the search format: instead of issuing proper search calls, the model relies on exposed trajectory content. Since this tool-use behavior is not strongly present in the base model, the policy has less prior capability to fall back on once active search is weakened.

Overall, these results suggest that the failure of full-expert-trace conditioning is not specific to one model, one expert source, or one domain, and that masking expert traces provides a more reliable way to use expert guidance in RLVR.

\begin{figure*}[t]
\centering
\small
\setlength{\tabcolsep}{5pt}

\begin{minipage}[b]{0.45\textwidth}
    \centering
    \includegraphics[width=0.88\linewidth]{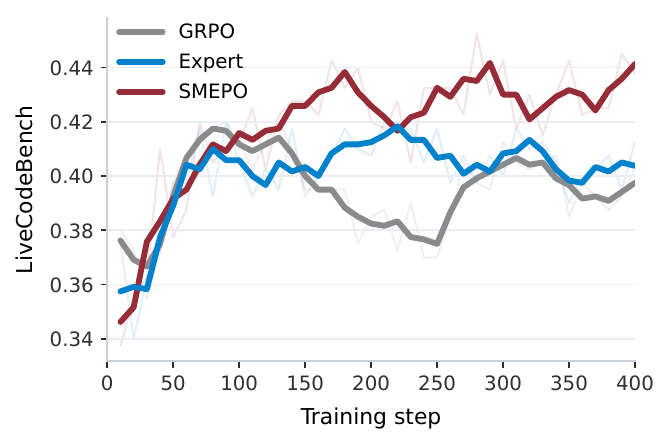}

    \vspace{0.25em}
    {\footnotesize (a) LiveCodeBench curve.}
\end{minipage}
\hfill
\begin{minipage}[b]{0.50\textwidth}
    \centering
    \renewcommand{\arraystretch}{1.08}
    \begin{tabular}{llcc}
    \toprule
    Model & Method & Step & Time (hr) \\
    \midrule
    \multirow{2}{*}{Qwen3-8B-Base}
    & GRPO   & 255 & 9.90 \\
    & \cellcolor{mitblue}SMEPO & \cellcolor{mitblue}55  & \cellcolor{mitblue}2.37 \\
    \midrule
    \multirow{2}{*}{Qwen2.5-7B}
    & GRPO   & 300 & 8.00 \\
    & \cellcolor{mitblue}SMEPO & \cellcolor{mitblue}175 & \cellcolor{mitblue}4.57 \\
    \midrule
    \multirow{2}{*}{R1-Distill-Qwen-7B}
    & GRPO   & 350 & 14.56 \\
    & \cellcolor{mitblue}SMEPO & \cellcolor{mitblue}270 & \cellcolor{mitblue}10.99 \\
    \bottomrule
    \end{tabular}

    \vspace{0.25em}
    {\footnotesize (b) Steps and wall-clock time to target.}
\end{minipage}

\caption{\textbf{Training dynamics and target-reaching efficiency.}
(Left): LiveCodeBench curve for Qwen3-8B-Base with DeepSeek-R1 traces. (Right): time to reach the GRPO peak mean@32 accuracy (averaged over AIME25 and AIME26).}
\label{fig:efficiency}
\end{figure*}

\subsection{In-depth analysis beyond accuracy}

\paragraph{Training efficiency.}
Figure~\ref{fig:efficiency}(b) reports the time required to reach a fixed hard-math target. SMEPO reaches this target significantly faster than vanilla GRPO across all three models. For instance, on Qwen3-8B-Base, it requires only 2.37 hours compared to GRPO’s 9.90 hours. Consequently, SMEPO enhances not only the final accuracy but also the convergence efficiency.

\paragraph{Analysis of copying behavior.} 
Figure~\ref{fig:code-copy-diagnostics} illustrates two copying patterns. Unlike full-expert conditioning, which yields high \emph{visible-trace overlap} and \emph{expert-code similarity}, SMEPO maintains near-zero overlap by masking executable solution spans. This demonstrates that complete code traces create an unintended reward channel, where policies gain rewards by simply reproducing prompt-exposed content. By suppressing this channel, SMEPO ensures the model utilizes high-level guidance—such as algorithmic plans and task interpretation—rather than merely adapting expert code. See Appendix~\ref{app:code-copy} for detailed metric definitions.

\begin{table*}[t]
\centering
\small
\setlength{\tabcolsep}{6pt}
\begin{tabular}{lccccccc}
\toprule
Method & GSM8K & MATH500 & AIME25 & AIME26 & AMC & Olympiad & Avg. \\
\midrule
Qwen3-8B-Base              & 68.0 & 67.0 & 9.9 & 7.6 & 50.0 & \underline{25.8} & 38.1\\
GHPO        & \underline{78.3} & \underline{72.8} & \underline{13.4} & \textbf{10.2} & \underline{60.0} & \textbf{29.5} & \underline{44.0} \\
\rowcolor{mitblue}
GHPO + SMEPO & \textbf{82.1} & \textbf{74.2} & \textbf{15.1} & \underline{10.1} & \textbf{62.5} & \textbf{29.5} & \textbf{45.6} \\
\bottomrule
\end{tabular}
\caption{\textbf{Additional gains with GHPO.}
On math benchmarks with Qwen3-8B-Base, adding SMEPO to GHPO improves the six-benchmark average from 44.0 to 45.6, suggesting that masked expert traces can bring additional gains on top of prompt-refinement guidance. Best results are in bold; second-best results are underlined.}
\label{tab:ghpo}
\end{table*}

\subsection{Compatibility with other guidance methods}

\begin{figure*}[t]
\centering
\begin{subfigure}[t]{0.42\textwidth}
    \centering
    \includegraphics[width=\linewidth]{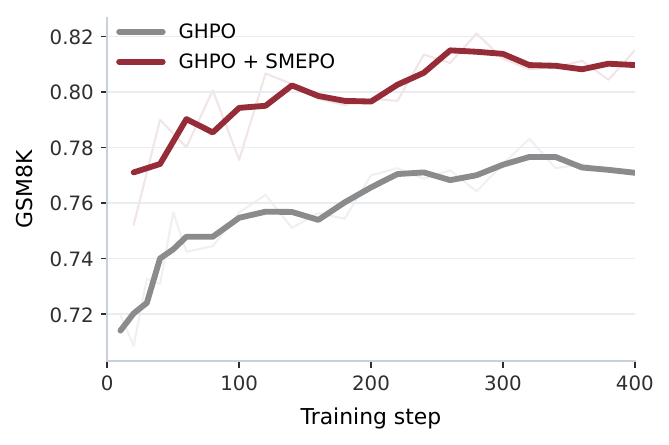}
    \caption{GSM8K.}
\end{subfigure}\hfill
\begin{subfigure}[t]{0.42\textwidth}
    \centering
    \includegraphics[width=\linewidth]{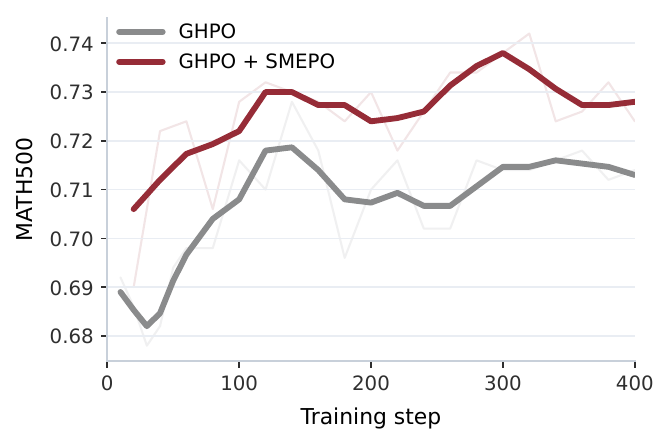}
    \caption{MATH500.}
\end{subfigure}
\caption{\textbf{Training curves with GHPO.}
Using the original GHPO setup, adding SMEPO improves the model accuracy on GSM8K and MATH500.}
\label{fig:ghpo-curves}
\end{figure*}

Table~\ref{tab:ghpo} evaluates SMEPO integrated with GHPO using identical training configurations. Adding SMEPO masking improves Qwen3-8B-Base math performance from 44.0 to 45.6. As shown in Figure~\ref{fig:ghpo-curves}, these gains are additive rather than redundant, suggesting masked traces provide unique benefits beyond GHPO's original signals.

\begin{table*}[t]
\centering
\small
\setlength{\tabcolsep}{4.2pt}
\renewcommand{\arraystretch}{1.05}

\begin{tabularx}{\textwidth}{llYYYYYYY}
\toprule
Cat. & Method & GSM8K & MATH500 & AIME25 & AIME26 & AMC & Olympiad & Avg. \\
\midrule
Full
& Expert (100\%)  & 76.4 & 69.2 & 11.4 & 8.6 & 52.5 & 27.7 & 41.0 \\
\midrule
\multirow{3}{*}{Prefix}
& Expert (75\%)   & 86.1 & 75.0 & \underline{14.5} & 11.6 & 60.0 & 29.7 & 46.1 \\
& Expert (50\%)   & 85.2 & 72.6 & 13.5 & 10.3 & 57.5 & 28.8 & 44.7 \\
& Expert (25\%)   & 87.1 & 74.6 & 14.2 & 11.0 & \underline{65.0} & \underline{30.6} & 47.1 \\
\midrule
\multirow{2}{*}{Random}
& Word            & 73.6 & 67.0 & 10.2 & 8.4 & 50.0 & 25.2 & 39.1 \\
& Sentence        & 75.7 & 68.0 & 10.9 & 9.4 & 45.0 & 26.9 & 39.3 \\
\midrule
\multirow{3}{*}{Targeted}
& Mask Final      & 82.9 & 70.4 & 11.7 & 9.1 & 57.5 & 27.9 & 43.2 \\
& Mask Ans. Uses  & \underline{89.9} & \underline{75.2} & 14.4 & \underline{13.2} & \underline{65.0} & 30.4 & \underline{48.0} \\
& \cellcolor{mitblue}SMEPO
  & \cellcolor{mitblue} \textbf{92.2}
  & \cellcolor{mitblue} \textbf{76.6}
  & \cellcolor{mitblue} \textbf{17.6}
  & \cellcolor{mitblue} \textbf{13.4}
  & \cellcolor{mitblue} \textbf{70.0}
  & \cellcolor{mitblue} \textbf{32.0}
  & \cellcolor{mitblue} \textbf{50.3} \\
\bottomrule
\end{tabularx}

\caption{\textbf{Math ablations with Qwen3-8B-Base.}
We compare full-expert-trace, prefix exposure, random masking, and targeted masking. Best results are in bold; second-best results are underlined.}
\label{tab:ablation}
\end{table*}




\subsection{Ablations}
\label{sec:ablation}

Table~\ref{tab:ablation} studies which part of the expert-trace transformation is responsible for the gains. First, prefix exposure shows that simply revealing less of the trace is insufficient. Using 75\%, 50\%, or 25\% of the expert prefix improves over full-expert-trace conditioning, but the gains are non-monotonic and remain below SMEPO. This suggests that the issue is not only the length or amount of expert text, but whether reward-relevant spans remain visible.

Second, random masking performs poorly. Random word masking and random sentence masking both underperform full-expert-trace conditioning and are far below SMEPO. Both random baselines use a masking budget matched to SMEPO for each problem. Thus, the benefit of SMEPO is not due to the amount of removed text, generic noise injection, or prompt corruption.

Third, reward-aware masking shows a clear progression. Masking only the final answer improves over full-expert-trace conditioning; masking all answer occurrences improves further; and SMEPO performs best by masking numeric content throughout the math trace. This supports our diagnosis that reward-relevant information is distributed across expert traces: final answers, repeated answer values, and intermediate numeric values can all act as exploitable reward channels. SMEPO works best because it removes these channels while preserving the surrounding procedural guidance.

\section{Related Work}
\label{sec:related}

\paragraph{RLVR and guided reasoning.}
RLVR improves reasoning models with verifier-based feedback in math and code \citep{deepseekmath,deepseekr1,reinforcepp,dapo,oneshotrlvr}, and has been extended to agentic search and research environments with tool-use and retrieval trajectories, such as Search-R1 \citep{searchr1} and DeepResearcher \citep{deepresearcher}. Recent work studies how to scale RLVR, stabilize critic-free optimization, and improve exploration under sparse rewards \citep{dapo,understandr1zero,oneshotrlvr}. Closely related methods use off-policy or privileged guidance: LUFFY \citep{luffy} incorporates off-policy reasoning traces during RL; Guide \citep{guide} injects hints when all initial rollouts fail; GHPO \citep{ghpo} provides difficulty-aware prompt refinement; TRAPO \citep{trapo} interleaves expert-prefix supervision with RL; POPE \citep{pope} uses oracle prefixes for hard problems; and Learning-like-humans \citep{learninglikehumans} uses expert traces to reformulate hard problems. These methods mainly regulate \emph{when} or \emph{how much} expert signal is shown, while we ask \emph{which parts} of an expert trace should be shown.

\paragraph{Reward hacking under expert guidance.}
Reward hacking occurs when a learner obtains high reward through unintended strategies rather than the intended behavior \citep{amodei2016concrete,everitt2021reward,skalse2022defining}. We study a guidance-induced form, where expert traces expose reward-relevant values, executable programs, or entities along the critical path that the policy can exploit during RL.

\paragraph{Masked and fill-in-the-blank-style learning.}
Masking and infilling learn from partially observed context: BERT \citep{bert}, MASS \citep{mass}, BART \citep{bart}, and T5 \citep{t5} reconstruct masked or corrupted text, while fill-in-the-middle training \citep{fim} conditions on prefix and suffix context. Code models also use FIM or fill-in-the-blank objectives, including InCoder \citep{incoder}, SantaCoder \citep{santacoder}, CodeT5+ \citep{codet5plus}, DeepSeek-Coder \citep{deepseekcoder}, and DeepSeek-Coder-V2 \citep{deepseekcoderv2}, with syntax- or structure-aware variants such as SAFIM \citep{safim} and AST-FIM \citep{astfim}. SMEPO brings a related masking principle into expert-guided RL, using masked traces as exploration guidance rather than supervised reconstruction targets.

\section{Conclusion}
\label{sec:conclusion}

Expert traces provide useful guidance but can also expose reward-informative content, including final answers and critical-path intermediate information. SMEPO masks these spans while preserving procedural cues, turning expert traces into fill-in-the-blank-style guidance for RLVR. Across math, code, and agentic search, SMEPO improves over vanilla GRPO and full-expert-trace conditioning, suggesting that effective expert-guided RL should preserve useful guidance while masking reward-relevant content along the critical path, rather than exposing expert traces unchanged.

{
    \small
    \bibliographystyle{unsrtnat}
    \bibliography{references}

@article{deepseekr1,
   title={{DeepSeek-R1} incentivizes reasoning in {LLM}s through reinforcement learning},
   volume={645},
   ISSN={1476-4687},
   url={http://dx.doi.org/10.1038/s41586-025-09422-z},
   DOI={10.1038/s41586-025-09422-z},
   number={8081},
   journal={Nature},
   publisher={Springer Science and Business Media LLC},
   author={Guo, Daya and Yang, Dejian and Zhang, Haowei and Song, Junxiao and Wang, Peiyi and Zhu, Qihao and Xu, Runxin and Zhang, Ruoyu and Ma, Shirong and Bi, Xiao and Zhang, Xiaokang and Yu, Xingkai and Wu, Yu and Wu, Z. F. and Gou, Zhibin and Shao, Zhihong and Li, Zhuoshu and Gao, Ziyi and Liu, Aixin and Xue, Bing and Wang, Bingxuan and Wu, Bochao and Feng, Bei and Lu, Chengda and Zhao, Chenggang and Deng, Chengqi and Ruan, Chong and Dai, Damai and Chen, Deli and Ji, Dongjie and Li, Erhang and Lin, Fangyun and Dai, Fucong and Luo, Fuli and Hao, Guangbo and Chen, Guanting and Li, Guowei and Zhang, H. and Xu, Hanwei and Ding, Honghui and Gao, Huazuo and Qu, Hui and Li, Hui and Guo, Jianzhong and Li, Jiashi and Chen, Jingchang and Yuan, Jingyang and Tu, Jinhao and Qiu, Junjie and Li, Junlong and Cai, J. L. and Ni, Jiaqi and Liang, Jian and Chen, Jin and Dong, Kai and Hu, Kai and You, Kaichao and Gao, Kaige and Guan, Kang and Huang, Kexin and Yu, Kuai and Wang, Lean and Zhang, Lecong and Zhao, Liang and Wang, Litong and Zhang, Liyue and Xu, Lei and Xia, Leyi and Zhang, Mingchuan and Zhang, Minghua and Tang, Minghui and Zhou, Mingxu and Li, Meng and Wang, Miaojun and Li, Mingming and Tian, Ning and Huang, Panpan and Zhang, Peng and Wang, Qiancheng and Chen, Qinyu and Du, Qiushi and Ge, Ruiqi and Zhang, Ruisong and Pan, Ruizhe and Wang, Runji and Chen, R. J. and Jin, R. L. and Chen, Ruyi and Lu, Shanghao and Zhou, Shangyan and Chen, Shanhuang and Ye, Shengfeng and Wang, Shiyu and Yu, Shuiping and Zhou, Shunfeng and Pan, Shuting and Li, S. S. and Zhou, Shuang and Wu, Shaoqing and Yun, Tao and Pei, Tian and Sun, Tianyu and Wang, T. and Zeng, Wangding and Liu, Wen and Liang, Wenfeng and Gao, Wenjun and Yu, Wenqin and Zhang, Wentao and Xiao, W. L. and An, Wei and Liu, Xiaodong and Wang, Xiaohan and Chen, Xiaokang and Nie, Xiaotao and Cheng, Xin and Liu, Xin and Xie, Xin and Liu, Xingchao and Yang, Xinyu and Li, Xinyuan and Su, Xuecheng and Lin, Xuheng and Li, X. Q. and Jin, Xiangyue and Shen, Xiaojin and Chen, Xiaosha and Sun, Xiaowen and Wang, Xiaoxiang and Song, Xinnan and Zhou, Xinyi and Wang, Xianzu and Shan, Xinxia and Li, Y. K. and Wang, Y. Q. and Wei, Y. X. and Zhang, Yang and Xu, Yanhong and Li, Yao and Zhao, Yao and Sun, Yaofeng and Wang, Yaohui and Yu, Yi and Zhang, Yichao and Shi, Yifan and Xiong, Yiliang and He, Ying and Piao, Yishi and Wang, Yisong and Tan, Yixuan and Ma, Yiyang and Liu, Yiyuan and Guo, Yongqiang and Ou, Yuan and Wang, Yuduan and Gong, Yue and Zou, Yuheng and He, Yujia and Xiong, Yunfan and Luo, Yuxiang and You, Yuxiang and Liu, Yuxuan and Zhou, Yuyang and Zhu, Y. X. and Huang, Yanping and Li, Yaohui and Zheng, Yi and Zhu, Yuchen and Ma, Yunxian and Tang, Ying and Zha, Yukun and Yan, Yuting and Ren, Z. Z. and Ren, Zehui and Sha, Zhangli and Fu, Zhe and Xu, Zhean and Xie, Zhenda and Zhang, Zhengyan and Hao, Zhewen and Ma, Zhicheng and Yan, Zhigang and Wu, Zhiyu and Gu, Zihui and Zhu, Zijia and Liu, Zijun and Li, Zilin and Xie, Ziwei and Song, Ziyang and Pan, Zizheng and Huang, Zhen and Xu, Zhipeng and Zhang, Zhongyu and Zhang, Zhen},
   year={2025},
   month=sep, pages={633–638} }

@misc{openthoughts,
      title={{OpenThoughts}: Data Recipes for Reasoning Models}, 
      author={Etash Guha and Ryan Marten and Sedrick Keh and Negin Raoof and Georgios Smyrnis and Hritik Bansal and Marianna Nezhurina and Jean Mercat and Trung Vu and Zayne Sprague and Ashima Suvarna and Benjamin Feuer and Liangyu Chen and Zaid Khan and Eric Frankel and Sachin Grover and Caroline Choi and Niklas Muennighoff and Shiye Su and Wanjia Zhao and John Yang and Shreyas Pimpalgaonkar and Kartik Sharma and Charlie Cheng-Jie Ji and Yichuan Deng and Sarah Pratt and Vivek Ramanujan and Jon Saad-Falcon and Jeffrey Li and Achal Dave and Alon Albalak and Kushal Arora and Blake Wulfe and Chinmay Hegde and Greg Durrett and Sewoong Oh and Mohit Bansal and Saadia Gabriel and Aditya Grover and Kai-Wei Chang and Vaishaal Shankar and Aaron Gokaslan and Mike A. Merrill and Tatsunori Hashimoto and Yejin Choi and Jenia Jitsev and Reinhard Heckel and Maheswaran Sathiamoorthy and Alexandros G. Dimakis and Ludwig Schmidt},
      year={2025},
      eprint={2506.04178},
      archivePrefix={arXiv},
      primaryClass={cs.LG},
      url={https://arxiv.org/abs/2506.04178}, 
}

@misc{reinforcepp,
      title={{REINFORCE++}: Stabilizing Critic-Free Policy Optimization with Global Advantage Normalization}, 
      author={Jian Hu and Jason Klein Liu and Haotian Xu and Wei Shen},
      year={2025},
      eprint={2501.03262},
      archivePrefix={arXiv},
      primaryClass={cs.CL},
      url={https://arxiv.org/abs/2501.03262}, 
}

@misc{ghpo,
      title={{GHPO}: Adaptive Guidance for Stable and Efficient {LLM} Reinforcement Learning}, 
      author={Ziru Liu and Cheng Gong and Xinyu Fu and Yaofang Liu and Ran Chen and Shoubo Hu and Suiyun Zhang and Rui Liu and Qingfu Zhang and Dandan Tu},
      year={2025},
      eprint={2507.10628},
      archivePrefix={arXiv},
      primaryClass={cs.LG},
      url={https://arxiv.org/abs/2507.10628}, 
}

@misc{luffy,
      title={Learning to Reason under Off-Policy Guidance}, 
      author={Jianhao Yan and Yafu Li and Zican Hu and Zhi Wang and Ganqu Cui and Xiaoye Qu and Yu Cheng and Yue Zhang},
      year={2025},
      eprint={2504.14945},
      archivePrefix={arXiv},
      primaryClass={cs.LG},
      url={https://arxiv.org/abs/2504.14945}, 
}

@misc{guide,
      title={Adaptive Guidance Accelerates Reinforcement Learning of Reasoning Models}, 
      author={Vaskar Nath and Elaine Lau and Anisha Gunjal and Manasi Sharma and Nikhil Baharte and Sean Hendryx},
      year={2025},
      eprint={2506.13923},
      archivePrefix={arXiv},
      primaryClass={cs.LG},
      url={https://arxiv.org/abs/2506.13923}, 
}

@misc{trapo,
      title={Trust-Region Adaptive Policy Optimization}, 
      author={Mingyu Su and Jian Guan and Yuxian Gu and Minlie Huang and Hongning Wang},
      year={2025},
      eprint={2512.17636},
      archivePrefix={arXiv},
      primaryClass={cs.LG},
      url={https://arxiv.org/abs/2512.17636}, 
}

@misc{pope,
      title={{POPE}: Learning to Reason on Hard Problems via Privileged On-Policy Exploration}, 
      author={Yuxiao Qu and Amrith Setlur and Virginia Smith and Ruslan Salakhutdinov and Aviral Kumar},
      year={2026},
      eprint={2601.18779},
      archivePrefix={arXiv},
      primaryClass={cs.LG},
      url={https://arxiv.org/abs/2601.18779}, 
}

@software{spacy,
  author       = {Montani, Ines and Honnibal, Matthew and Boyd, Adriane and Van Landeghem, Sofie and Peters, Henning},
  title        = {explosion/{spaCy}: v3.7.2: Fixes for {APIs} and requirements},
  year         = {2023},
  month        = oct,
  publisher    = {Zenodo},
  version      = {v3.7.2},
  doi          = {10.5281/zenodo.10009823},
  url          = {https://doi.org/10.5281/zenodo.10009823},
}

@misc{gsm8k,
      title={Training Verifiers to Solve Math Word Problems}, 
      author={Karl Cobbe and Vineet Kosaraju and Mohammad Bavarian and Mark Chen and Heewoo Jun and Lukasz Kaiser and Matthias Plappert and Jerry Tworek and Jacob Hilton and Reiichiro Nakano and Christopher Hesse and John Schulman},
      year={2021},
      eprint={2110.14168},
      archivePrefix={arXiv},
      primaryClass={cs.LG},
      url={https://arxiv.org/abs/2110.14168}, 
}

@misc{olympiadbench,
      title={{OlympiadBench}: A Challenging Benchmark for Promoting AGI with Olympiad-Level Bilingual Multimodal Scientific Problems}, 
      author={Chaoqun He and Renjie Luo and Yuzhuo Bai and Shengding Hu and Zhen Leng Thai and Junhao Shen and Jinyi Hu and Xu Han and Yujie Huang and Yuxiang Zhang and Jie Liu and Lei Qi and Zhiyuan Liu and Maosong Sun},
      year={2024},
      eprint={2402.14008},
      archivePrefix={arXiv},
      primaryClass={cs.CL},
      url={https://arxiv.org/abs/2402.14008}, 
}

@misc{aime25,
      title={American Invitational Mathematics Examination ({AIME}) 2025}, 
      author={Zhang, Yifan and Math-AI, Team},
      year={2025},
      url={https://huggingface.co/datasets/math-ai/aime25},
}

@misc{aime26,
      title={American Invitational Mathematics Examination ({AIME}) 2026}, 
      author={Zhang, Yifan and Math-AI, Team},
      year={2026},
      url={https://huggingface.co/datasets/math-ai/aime26},
}

@misc{amc23,
      title={American Mathematics Competitions ({AMC}) 2023},
      author={Zhang, Yifan and Math-AI, Team},
      year={2023},
      url={https://huggingface.co/datasets/math-ai/amc23},
}

@misc{math500,
      title={Let's Verify Step by Step}, 
      author={Hunter Lightman and Vineet Kosaraju and Yura Burda and Harri Edwards and Bowen Baker and Teddy Lee and Jan Leike and John Schulman and Ilya Sutskever and Karl Cobbe},
      year={2023},
      eprint={2305.20050},
      archivePrefix={arXiv},
      primaryClass={cs.LG},
      url={https://arxiv.org/abs/2305.20050}, 
}

@misc{humaneval,
      title={Evaluating Large Language Models Trained on Code}, 
      author={Mark Chen and Jerry Tworek and Heewoo Jun and Qiming Yuan and Henrique Ponde de Oliveira Pinto and Jared Kaplan and Harri Edwards and Yuri Burda and Nicholas Joseph and Greg Brockman and Alex Ray and Raul Puri and Gretchen Krueger and Michael Petrov and Heidy Khlaaf and Girish Sastry and Pamela Mishkin and Brooke Chan and Scott Gray and Nick Ryder and Mikhail Pavlov and Alethea Power and Lukasz Kaiser and Mohammad Bavarian and Clemens Winter and Philippe Tillet and Felipe Petroski Such and Dave Cummings and Matthias Plappert and Fotios Chantzis and Elizabeth Barnes and Ariel Herbert-Voss and William Hebgen Guss and Alex Nichol and Alex Paino and Nikolas Tezak and Jie Tang and Igor Babuschkin and Suchir Balaji and Shantanu Jain and William Saunders and Christopher Hesse and Andrew N. Carr and Jan Leike and Josh Achiam and Vedant Misra and Evan Morikawa and Alec Radford and Matthew Knight and Miles Brundage and Mira Murati and Katie Mayer and Peter Welinder and Bob McGrew and Dario Amodei and Sam McCandlish and Ilya Sutskever and Wojciech Zaremba},
      year={2021},
      eprint={2107.03374},
      archivePrefix={arXiv},
      primaryClass={cs.LG},
      url={https://arxiv.org/abs/2107.03374}, 
}

@misc{humanevalplus,
      title={Is Your Code Generated by {ChatGPT} Really Correct? Rigorous Evaluation of Large Language Models for Code Generation}, 
      author={Jiawei Liu and Chunqiu Steven Xia and Yuyao Wang and Lingming Zhang},
      year={2023},
      eprint={2305.01210},
      archivePrefix={arXiv},
      primaryClass={cs.SE},
      url={https://arxiv.org/abs/2305.01210}, 
}

@misc{lcb,
      title={{LiveCodeBench}: Holistic and Contamination Free Evaluation of Large Language Models for Code}, 
      author={Naman Jain and King Han and Alex Gu and Wen-Ding Li and Fanjia Yan and Tianjun Zhang and Sida Wang and Armando Solar-Lezama and Koushik Sen and Ion Stoica},
      year={2024},
      eprint={2403.07974},
      archivePrefix={arXiv},
      primaryClass={cs.SE},
      url={https://arxiv.org/abs/2403.07974}, 
}

@misc{bamboogle,
      title={Measuring and Narrowing the Compositionality Gap in Language Models}, 
      author={Ofir Press and Muru Zhang and Sewon Min and Ludwig Schmidt and Noah A. Smith and Mike Lewis},
      year={2023},
      eprint={2210.03350},
      archivePrefix={arXiv},
      primaryClass={cs.CL},
      url={https://arxiv.org/abs/2210.03350}, 
}

@article{mmlu_1,
      title={Measuring Massive Multitask Language Understanding},
      author={Dan Hendrycks and Collin Burns and Steven Basart and Andy Zou and Mantas Mazeika and Dawn Song and Jacob Steinhardt},
      journal={Proceedings of the International Conference on Learning Representations (ICLR)},
      year={2021}
    }

@article{mmlu_2,
      title={Aligning AI With Shared Human Values},
      author={Dan Hendrycks and Collin Burns and Steven Basart and Andrew Critch and Jerry Li and Dawn Song and Jacob Steinhardt},
      journal={Proceedings of the International Conference on Learning Representations (ICLR)},
      year={2021}
    }

@misc{miroverse,
      title={{MiroThinker}: Pushing the Performance Boundaries of Open-Source Research Agents via Model, Context, and Interactive Scaling}, 
      author={MiroMind Team and Song Bai and Lidong Bing and Carson Chen and Guanzheng Chen and Yuntao Chen and Zhe Chen and Ziyi Chen and Jifeng Dai and Xuan Dong and Wenhan Dou and Yue Deng and Yunjie Fu and Junqi Ge and Chenxia Han and Tammy Huang and Zhenhang Huang and Jerry Jiao and Shilei Jiang and Tianyu Jiao and Xiaoqi Jian and Lei Lei and Ruilin Li and Gen Luo and Tiantong Li and Xiang Lin and Ziyuan Liu and Zhiqi Li and Jie Ni and Qiang Ren and Pax Sun and Shiqian Su and Chenxin Tao and Bin Wang and Wenhai Wang and Haonan Wang and James Wang and Jin Wang and Jojo Wang and Letian Wang and Shizun Wang and Weizhi Wang and Zixuan Wang and Jinfan Xu and Sen Xing and Chenyu Yang and Hai Ye and Jiaheng Yu and Yue Yu and Muyan Zhong and Tianchen Zhao and Xizhou Zhu and Yanpeng Zhou and Yifan Zhang and Zhi Zhu},
      year={2026},
      eprint={2511.11793},
      archivePrefix={arXiv},
      primaryClass={cs.CL},
      url={https://arxiv.org/abs/2511.11793}, 
}

@misc{deepseekmath,
      title={{DeepSeekMath}: Pushing the Limits of Mathematical Reasoning in Open Language Models}, 
      author={Zhihong Shao and Peiyi Wang and Qihao Zhu and Runxin Xu and Junxiao Song and Xiao Bi and Haowei Zhang and Mingchuan Zhang and Y. K. Li and Y. Wu and Daya Guo},
      year={2024},
      eprint={2402.03300},
      archivePrefix={arXiv},
      primaryClass={cs.CL},
      url={https://arxiv.org/abs/2402.03300}, 
}

@misc{oneshotrlvr,
      title={Reinforcement Learning for Reasoning in Large Language Models with One Training Example}, 
      author={Yiping Wang and Qing Yang and Zhiyuan Zeng and Liliang Ren and Liyuan Liu and Baolin Peng and Hao Cheng and Xuehai He and Kuan Wang and Jianfeng Gao and Weizhu Chen and Shuohang Wang and Simon Shaolei Du and Yelong Shen},
      year={2025},
      eprint={2504.20571},
      archivePrefix={arXiv},
      primaryClass={cs.LG},
      url={https://arxiv.org/abs/2504.20571}, 
}

@misc{amodei2016concrete,
      title={Concrete Problems in {AI} Safety}, 
      author={Dario Amodei and Chris Olah and Jacob Steinhardt and Paul Christiano and John Schulman and Dan Mané},
      year={2016},
      eprint={1606.06565},
      archivePrefix={arXiv},
      primaryClass={cs.AI},
      url={https://arxiv.org/abs/1606.06565}, 
}

@inproceedings{bert,
    title = "{BERT}: Pre-training of Deep Bidirectional Transformers for Language Understanding",
    author = "Devlin, Jacob  and
      Chang, Ming-Wei  and
      Lee, Kenton  and
      Toutanova, Kristina",
    editor = "Burstein, Jill  and
      Doran, Christy  and
      Solorio, Thamar",
    booktitle = "Proceedings of the 2019 Conference of the North {A}merican Chapter of the Association for Computational Linguistics: Human Language Technologies, Volume 1 (Long and Short Papers)",
    month = jun,
    year = "2019",
    address = "Minneapolis, Minnesota",
    publisher = "Association for Computational Linguistics",
    url = "https://aclanthology.org/N19-1423/",
    doi = "10.18653/v1/N19-1423",
    pages = "4171--4186"
}

@misc{fim,
      title={Efficient Training of Language Models to Fill in the Middle}, 
      author={Mohammad Bavarian and Heewoo Jun and Nikolas Tezak and John Schulman and Christine McLeavey and Jerry Tworek and Mark Chen},
      year={2022},
      eprint={2207.14255},
      archivePrefix={arXiv},
      primaryClass={cs.CL},
      url={https://arxiv.org/abs/2207.14255}, 
}

@misc{deepseekcoder,
      title={{DeepSeek-Coder}: When the Large Language Model Meets Programming -- The Rise of Code Intelligence}, 
      author={Daya Guo and Qihao Zhu and Dejian Yang and Zhenda Xie and Kai Dong and Wentao Zhang and Guanting Chen and Xiao Bi and Y. Wu and Y. K. Li and Fuli Luo and Yingfei Xiong and Wenfeng Liang},
      year={2024},
      eprint={2401.14196},
      archivePrefix={arXiv},
      primaryClass={cs.SE},
      url={https://arxiv.org/abs/2401.14196}, 
}

@misc{deepseekcoderv2,
      title={{DeepSeek-Coder-V2}: Breaking the Barrier of Closed-Source Models in Code Intelligence}, 
      author={{DeepSeek-AI} and Qihao Zhu and Daya Guo and Zhihong Shao and Dejian Yang and Peiyi Wang and Runxin Xu and Y. Wu and Yukun Li and Huazuo Gao and Shirong Ma and Wangding Zeng and Xiao Bi and Zihui Gu and Hanwei Xu and Damai Dai and Kai Dong and Liyue Zhang and Yishi Piao and Zhibin Gou and Zhenda Xie and Zhewen Hao and Bingxuan Wang and Junxiao Song and Deli Chen and Xin Xie and Kang Guan and Yuxiang You and Aixin Liu and Qiushi Du and Wenjun Gao and Xuan Lu and Qinyu Chen and Yaohui Wang and Chengqi Deng and Jiashi Li and Chenggang Zhao and Chong Ruan and Fuli Luo and Wenfeng Liang},
      year={2024},
      eprint={2406.11931},
      archivePrefix={arXiv},
      primaryClass={cs.SE},
      url={https://arxiv.org/abs/2406.11931}, 
}

@misc{qwen3technicalreport,
      title={Qwen3 Technical Report}, 
      author={An Yang and Anfeng Li and Baosong Yang and Beichen Zhang and Binyuan Hui and Bo Zheng and Bowen Yu and Chang Gao and Chengen Huang and Chenxu Lv and Chujie Zheng and Dayiheng Liu and Fan Zhou and Fei Huang and Feng Hu and Hao Ge and Haoran Wei and Huan Lin and Jialong Tang and Jian Yang and Jianhong Tu and Jianwei Zhang and Jianxin Yang and Jiaxi Yang and Jing Zhou and Jingren Zhou and Junyang Lin and Kai Dang and Keqin Bao and Kexin Yang and Le Yu and Lianghao Deng and Mei Li and Mingfeng Xue and Mingze Li and Pei Zhang and Peng Wang and Qin Zhu and Rui Men and Ruize Gao and Shixuan Liu and Shuang Luo and Tianhao Li and Tianyi Tang and Wenbiao Yin and Xingzhang Ren and Xinyu Wang and Xinyu Zhang and Xuancheng Ren and Yang Fan and Yang Su and Yichang Zhang and Yinger Zhang and Yu Wan and Yuqiong Liu and Zekun Wang and Zeyu Cui and Zhenru Zhang and Zhipeng Zhou and Zihan Qiu},
      year={2025},
      eprint={2505.09388},
      archivePrefix={arXiv},
      primaryClass={cs.CL},
      url={https://arxiv.org/abs/2505.09388}, 
}

@misc{qwen25technicalreport,
      title={Qwen2.5 Technical Report}, 
      author={{Qwen Team} and An Yang and Baosong Yang and Beichen Zhang and Binyuan Hui and Bo Zheng and Bowen Yu and Chengyuan Li and Dayiheng Liu and Fei Huang and Haoran Wei and Huan Lin and Jian Yang and Jianhong Tu and Jianwei Zhang and Jianxin Yang and Jiaxi Yang and Jingren Zhou and Junyang Lin and Kai Dang and Keming Lu and Keqin Bao and Kexin Yang and Le Yu and Mei Li and Mingfeng Xue and Pei Zhang and Qin Zhu and Rui Men and Runji Lin and Tianhao Li and Tianyi Tang and Tingyu Xia and Xingzhang Ren and Xuancheng Ren and Yang Fan and Yang Su and Yichang Zhang and Yu Wan and Yuqiong Liu and Zeyu Cui and Zhenru Zhang and Zihan Qiu},
      year={2025},
      eprint={2412.15115},
      archivePrefix={arXiv},
      primaryClass={cs.CL},
      url={https://arxiv.org/abs/2412.15115}, 
}

@misc{dapo,
      title={{DAPO}: An Open-Source {LLM} Reinforcement Learning System at Scale}, 
      author={Qiying Yu and Zheng Zhang and Ruofei Zhu and Yufeng Yuan and Xiaochen Zuo and Yu Yue and Weinan Dai and Tiantian Fan and Gaohong Liu and Lingjun Liu and Xin Liu and Haibin Lin and Zhiqi Lin and Bole Ma and Guangming Sheng and Yuxuan Tong and Chi Zhang and Mofan Zhang and Wang Zhang and Hang Zhu and Jinhua Zhu and Jiaze Chen and Jiangjie Chen and Chengyi Wang and Hongli Yu and Yuxuan Song and Xiangpeng Wei and Hao Zhou and Jingjing Liu and Wei-Ying Ma and Ya-Qin Zhang and Lin Yan and Mu Qiao and Yonghui Wu and Mingxuan Wang},
      year={2025},
      eprint={2503.14476},
      archivePrefix={arXiv},
      primaryClass={cs.LG},
      url={https://arxiv.org/abs/2503.14476}, 
}

@misc{understandr1zero,
      title={Understanding {R1-Zero-Like} Training: A Critical Perspective}, 
      author={Zichen Liu and Changyu Chen and Wenjun Li and Penghui Qi and Tianyu Pang and Chao Du and Wee Sun Lee and Min Lin},
      year={2025},
      eprint={2503.20783},
      archivePrefix={arXiv},
      primaryClass={cs.LG},
      url={https://arxiv.org/abs/2503.20783}, 
}

@misc{deepresearcher,
      title={{DeepResearcher}: Scaling Deep Research via Reinforcement Learning in Real-world Environments}, 
      author={Yuxiang Zheng and Dayuan Fu and Xiangkun Hu and Xiaojie Cai and Lyumanshan Ye and Pengrui Lu and Pengfei Liu},
      year={2025},
      eprint={2504.03160},
      archivePrefix={arXiv},
      primaryClass={cs.AI},
      url={https://arxiv.org/abs/2504.03160}, 
}

@misc{searchr1,
      title={{Search-R1}: Training {LLMs} to Reason and Leverage Search Engines with Reinforcement Learning}, 
      author={Bowen Jin and Hansi Zeng and Zhenrui Yue and Jinsung Yoon and Sercan Arik and Dong Wang and Hamed Zamani and Jiawei Han},
      year={2025},
      eprint={2503.09516},
      archivePrefix={arXiv},
      primaryClass={cs.CL},
      url={https://arxiv.org/abs/2503.09516}, 
}

@misc{everitt2021reward,
      title={Reward Tampering Problems and Solutions in Reinforcement Learning: A Causal Influence Diagram Perspective}, 
      author={Tom Everitt and Marcus Hutter and Ramana Kumar and Victoria Krakovna},
      year={2021},
      eprint={1908.04734},
      archivePrefix={arXiv},
      primaryClass={cs.AI},
      url={https://arxiv.org/abs/1908.04734}, 
}

@inproceedings{skalse2022defining,
 author = {Skalse, Joar and Howe, Nikolaus and Krasheninnikov, Dmitrii and Krueger, David},
 booktitle = {Advances in Neural Information Processing Systems},
 editor = {S. Koyejo and S. Mohamed and A. Agarwal and D. Belgrave and K. Cho and A. Oh},
 pages = {9460--9471},
 publisher = {Curran Associates, Inc.},
 title = {Defining and Characterizing Reward Gaming},
 url = {https://proceedings.neurips.cc/paper_files/paper/2022/file/3d719fee332caa23d5038b8a90e81796-Paper-Conference.pdf},
 volume = {35},
 year = {2022}
}

@misc{mass,
      title={{MASS}: Masked Sequence to Sequence Pre-training for Language Generation}, 
      author={Kaitao Song and Xu Tan and Tao Qin and Jianfeng Lu and Tie-Yan Liu},
      year={2019},
      eprint={1905.02450},
      archivePrefix={arXiv},
      primaryClass={cs.CL},
      url={https://arxiv.org/abs/1905.02450}, 
}

@inproceedings{bart,
    title = "{BART}: Denoising Sequence-to-Sequence Pre-training for Natural Language Generation, Translation, and Comprehension",
    author = "Lewis, Mike  and
      Liu, Yinhan  and
      Goyal, Naman  and
      Ghazvininejad, Marjan  and
      Mohamed, Abdelrahman  and
      Levy, Omer  and
      Stoyanov, Veselin  and
      Zettlemoyer, Luke",
    editor = "Jurafsky, Dan  and
      Chai, Joyce  and
      Schluter, Natalie  and
      Tetreault, Joel",
    booktitle = "Proceedings of the 58th Annual Meeting of the Association for Computational Linguistics",
    month = jul,
    year = "2020",
    address = "Online",
    publisher = "Association for Computational Linguistics",
    url = "https://aclanthology.org/2020.acl-main.703/",
    doi = "10.18653/v1/2020.acl-main.703",
    pages = "7871--7880"
}

@article{t5,
  author  = {Colin Raffel and Noam Shazeer and Adam Roberts and Katherine Lee and Sharan Narang and Michael Matena and Yanqi Zhou and Wei Li and Peter J. Liu},
  title   = {Exploring the Limits of Transfer Learning with a Unified Text-to-Text Transformer},
  journal = {Journal of Machine Learning Research},
  year    = {2020},
  volume  = {21},
  number  = {140},
  pages   = {1-67},
  url     = {http://jmlr.org/papers/v21/20-074.html}
}

@misc{incoder,
      title={{InCoder}: A Generative Model for Code Infilling and Synthesis}, 
      author={Daniel Fried and Armen Aghajanyan and Jessy Lin and Sida Wang and Eric Wallace and Freda Shi and Ruiqi Zhong and Wen-tau Yih and Luke Zettlemoyer and Mike Lewis},
      year={2023},
      eprint={2204.05999},
      archivePrefix={arXiv},
      primaryClass={cs.SE},
      url={https://arxiv.org/abs/2204.05999}, 
}

@misc{santacoder,
      title={{SantaCoder}: don't reach for the stars!}, 
      author={Loubna Ben Allal and Raymond Li and Denis Kocetkov and Chenghao Mou and Christopher Akiki and Carlos Munoz Ferrandis and Niklas Muennighoff and Mayank Mishra and Alex Gu and Manan Dey and Logesh Kumar Umapathi and Carolyn Jane Anderson and Yangtian Zi and Joel Lamy Poirier and Hailey Schoelkopf and Sergey Troshin and Dmitry Abulkhanov and Manuel Romero and Michael Lappert and Francesco De Toni and Bernardo García del Río and Qian Liu and Shamik Bose and Urvashi Bhattacharyya and Terry Yue Zhuo and Ian Yu and Paulo Villegas and Marco Zocca and Sourab Mangrulkar and David Lansky and Huu Nguyen and Danish Contractor and Luis Villa and Jia Li and Dzmitry Bahdanau and Yacine Jernite and Sean Hughes and Daniel Fried and Arjun Guha and Harm de Vries and Leandro von Werra},
      year={2023},
      eprint={2301.03988},
      archivePrefix={arXiv},
      primaryClass={cs.SE},
      url={https://arxiv.org/abs/2301.03988}, 
}

@misc{codet5plus,
      title={{CodeT5+}: Open Code Large Language Models for Code Understanding and Generation}, 
      author={Yue Wang and Hung Le and Akhilesh Deepak Gotmare and Nghi D. Q. Bui and Junnan Li and Steven C. H. Hoi},
      year={2023},
      eprint={2305.07922},
      archivePrefix={arXiv},
      primaryClass={cs.CL},
      url={https://arxiv.org/abs/2305.07922}, 
}

@misc{safim,
      title={Evaluation of {LLMs} on Syntax-Aware Code Fill-in-the-Middle Tasks}, 
      author={Linyuan Gong and Sida Wang and Mostafa Elhoushi and Alvin Cheung},
      year={2024},
      eprint={2403.04814},
      archivePrefix={arXiv},
      primaryClass={cs.CL},
      url={https://arxiv.org/abs/2403.04814}, 
}

@misc{astfim,
      title={Structure-Aware Fill-in-the-Middle Pretraining for Code}, 
      author={Linyuan Gong and Alvin Cheung and Mostafa Elhoushi and Sida Wang},
      year={2025},
      eprint={2506.00204},
      archivePrefix={arXiv},
      primaryClass={cs.CL},
      url={https://arxiv.org/abs/2506.00204}, 
}

@inproceedings{verl, series={EuroSys ’25},
   title={{HybridFlow}: A Flexible and Efficient {RLHF} Framework},
   url={http://dx.doi.org/10.1145/3689031.3696075},
   DOI={10.1145/3689031.3696075},
   booktitle={Proceedings of the Twentieth European Conference on Computer Systems},
   publisher={ACM},
   author={Sheng, Guangming and Zhang, Chi and Ye, Zilingfeng and Wu, Xibin and Zhang, Wang and Zhang, Ru and Peng, Yanghua and Lin, Haibin and Wu, Chuan},
   year={2025},
   month=mar, pages={1279–1297},
   collection={EuroSys ’25} }

@misc{eval-harness,
  author       = {Gao, Leo and Tow, Jonathan and Abbasi, Baber and Biderman, Stella and Black, Sid and DiPofi, Anthony and Foster, Charles and Golding, Laurence and Hsu, Jeffrey and Le Noac'h, Alain and Li, Haonan and McDonell, Kyle and Muennighoff, Niklas and Ociepa, Chris and Phang, Jason and Reynolds, Laria and Schoelkopf, Hailey and Skowron, Aviya and Sutawika, Lintang and Tang, Eric and Thite, Anish and Wang, Ben and Wang, Kevin and Zou, Andy},
  title        = {The Language Model Evaluation Harness},
  month        = jul,
  year         = 2024,
  publisher    = {Zenodo},
  version      = {v0.4.3},
  doi          = {10.5281/zenodo.12608602},
  url          = {https://zenodo.org/records/12608602}
}

@misc{learninglikehumans,
      title={{Learning Like Humans}: Advancing {LLM} Reasoning Capabilities via Adaptive Difficulty Curriculum Learning and Expert-Guided Self-Reformulation}, 
      author={Enci Zhang and Xingang Yan and Wei Lin and Tianxiang Zhang and Qianchun Lu},
      year={2025},
      eprint={2505.08364},
      archivePrefix={arXiv},
      primaryClass={cs.AI},
      url={https://arxiv.org/abs/2505.08364}, 
}

@inproceedings{gliner,
    title = "{GL}i{NER}: Generalist Model for Named Entity Recognition using Bidirectional Transformer",
    author = "Zaratiana, Urchade  and
      Tomeh, Nadi  and
      Holat, Pierre  and
      Charnois, Thierry",
    editor = "Duh, Kevin  and
      Gomez, Helena  and
      Bethard, Steven",
    booktitle = "Proceedings of the 2024 Conference of the North American Chapter of the Association for Computational Linguistics: Human Language Technologies (Volume 1: Long Papers)",
    month = jun,
    year = "2024",
    address = "Mexico City, Mexico",
    publisher = "Association for Computational Linguistics",
    url = "https://aclanthology.org/2024.naacl-long.300/",
    doi = "10.18653/v1/2024.naacl-long.300",
    pages = "5364--5376"
}

@misc{sparse,
      title={Learn Hard Problems During {RL} with Reference Guided Fine-tuning}, 
      author={Yangzhen Wu and Shanda Li and Zixin Wen and Xin Zhou and Ameet Talwalkar and Yiming Yang and Wenhao Huang and Tianle Cai},
      year={2026},
      eprint={2603.01223},
      archivePrefix={arXiv},
      primaryClass={cs.LG},
      url={https://arxiv.org/abs/2603.01223}, 
}

@misc{reft,
      title={{ReFT}: Reasoning with Reinforced Fine-Tuning}, 
      author={Trung Quoc Luong and Xinbo Zhang and Zhanming Jie and Peng Sun and Xiaoran Jin and Hang Li},
      year={2024},
      eprint={2401.08967},
      archivePrefix={arXiv},
      primaryClass={cs.CL},
      url={https://arxiv.org/abs/2401.08967}, 
}
}

\newpage
\appendix

\section{Additional Results}
\label{app:additional-results}

\subsection{Additional LiveCodeBench v6 subset evaluation}
\label{app:lcb-v6}

We further evaluate Qwen3-8B-Base on a commonly used LiveCodeBench v6 subset, consisting of 131 problems released between 2025-02-01 and 2025-05-01. This evaluation complements the LiveCodeBench result in the main code table and provides an additional check on a recent code-generation subset used by the community.

\begin{figure}[t]
    \centering
    \begin{minipage}[c]{0.58\textwidth}
        \centering
        \includegraphics[width=\linewidth]{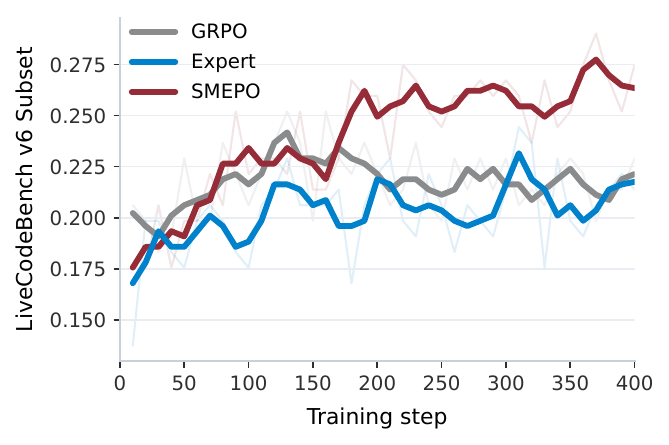}
    \end{minipage}
    \hfill
    \begin{minipage}[c]{0.38\textwidth}
        \centering
        \small
        \setlength{\tabcolsep}{6pt}
        \renewcommand{\arraystretch}{1.05}
        \begin{tabular}{lc}
        \toprule
        \rowcolor{blue!22}
        \multicolumn{2}{c}{\textbf{LCB v6 Subset}} \\
        \midrule
        Method & Pass@1 \\
        \midrule
        Qwen3-8B-Base & 14.5 \\
        GRPO          & \underline{25.2} \\
        Expert        & 24.4 \\
        \rowcolor{mitblue}
        SMEPO         & \textbf{29.0} \\
        \bottomrule
        \end{tabular}
    \end{minipage}

\caption{\textbf{Additional LiveCodeBench v6 subset evaluation.}
Left: Training dynamics; right: final pass@1 scores.
We evaluate Qwen3-8B-Base on 131 problems released between 2025-02-01 and 2025-05-01.
SMEPO performs best, consistent with the main code evaluation.}
\label{fig:lcb-v6-subset}
\end{figure}

Figure~\ref{fig:lcb-v6-subset} reports both the training dynamics and final pass@1 scores on this subset. SMEPO achieves the best final performance, improving over GRPO from 25.2 to 29.0 and over full expert-trace conditioning from 24.4 to 29.0. The result is consistent with our main code evaluation: directly exposing complete expert code can hurt performance relative to vanilla GRPO, while masking executable solution content yields better performance.

\subsection{Large-scale 89k math pool}
\label{app:large-scale-89k}

We additionally evaluate SMEPO on a larger 89k-example math training pool with Qwen3-8B-Base to test whether the same trend holds beyond the curated 1k hard-reasoning setting.

Table~\ref{tab:qwen3-89k-math} shows a consistent pattern. Vanilla GRPO improves the six-benchmark average from 38.1 to 47.8, while direct expert conditioning with DeepSeek-R1 traces decreases the average to 36.6. In contrast, SMEPO achieves 49.4, outperforming both vanilla GRPO and full-expert-trace conditioning. This result suggests that the benefit of masking is not limited to the small curated training set and remains effective at a larger data scale.

\subsection{General-domain checks on MMLU}
\label{app:mmlu}

We report MMLU-Overall as a general-domain sanity check rather than a main target metric. Since SMEPO modifies the guidance context used during GRPO, one concern is that the gains on math, code, or agentic search might come at the cost of broader model capability. MMLU is not designed to measure the specific reasoning behaviors targeted by SMEPO, but it provides a useful check of whether training causes a broad regression outside the target tasks.

Table~\ref{tab:supp-mmlu} shows that SMEPO remains essentially flat relative to the base model and vanilla GRPO across the math, code, and agentic-search settings. In several cases, SMEPO slightly improves MMLU-Overall, while SFT more visibly lowers MMLU. This is consistent with the role of SMEPO as a guidance transformation rather than pure imitation: it changes what expert context the policy sees during GRPO, but it still optimizes the same verifier-based objective. Overall, the MMLU results suggest that the task gains from masked expert guidance do not come from sacrificing broad general-domain performance.

\subsection{Diagnostic summary of full-expert-trace conditioning}
\label{app:full-expert-trace-diagnosis}

Table~\ref{tab:full_trace_diagnosis} summarizes the effect of directly appending complete expert traces during GRPO. Across all reported settings, full-expert-trace conditioning underperforms vanilla GRPO, although the traces come from strong expert models such as DeepSeek-R1 or Gemini-3-Pro. This supports the central observation of the paper: complete expert traces are not simply helpful demonstrations, because they can expose content that is too directly tied to verifier reward.

The drop is largest in agentic search, where complete trajectories can expose answer-relevant entities and search pivots that suppress the model's active use of the search format. The degradation is smaller for code and for DeepSeek-R1-Distill-Qwen-7B, but the direction remains negative. This matches the analysis in the main text: code traces require reproducing or adapting a longer executable solution, and the reasoning-distilled model has stronger existing reasoning ability, so both settings are less vulnerable than base-model math or agentic search. Overall, this diagnostic table reinforces our interpretation that expert traces are useful only when their guidance is separated from directly reward-relevant content.

\begin{table*}[t]
\centering
\small
\setlength{\tabcolsep}{6pt}
\begin{tabular}{lccccccc}
\toprule
Method & GSM8K & MATH500 & AIME25 & AIME26 & AMC & Olympiad & Avg. \\
\midrule
Qwen3-8B-Base             & 68.0 & 67.0 & 9.9 & 7.6 & 50.0 & 25.8 & 38.1 \\
GRPO           & \underline{92.3} & \underline{77.4} & \underline{13.0} & \underline{9.3} & \underline{65.0} & \underline{30.1} & \underline{47.8} \\
Expert (DS)   & 75.7 & 63.2 & 9.0 & 6.9 & 42.5 & 22.7 & 36.6 \\
SMEPO (DS) & \textbf{93.6} & \textbf{77.6} & \textbf{14.7} & \textbf{11.4} & \textbf{67.5} & \textbf{31.5} & \textbf{49.4} \\
\bottomrule
\end{tabular}
\caption{\textbf{Large-scale 89k-example math results with Qwen3-8B-Base.}
The same pattern persists at scale: full expert-trace conditioning underperforms vanilla GRPO, while SMEPO improves beyond both. Best results are in bold; second-best results are underlined.}
\label{tab:qwen3-89k-math}
\end{table*}

\begin{table*}[t]
\centering
\small
\setlength{\tabcolsep}{5pt}
\renewcommand{\arraystretch}{1.05}

\begin{subtable}[t]{0.36\textwidth}
\centering
\caption*{\textbf{Qwen3-8B-Base Math}}
\begin{tabular}{lc}
\toprule
Method & MMLU \\
\midrule
Qwen3-8B-Base      & 75.7 \\
GRPO               & 75.7 \\
\midrule
SFT (DS)           & 74.5 \\
SFT (DS) + RL      & 74.8 \\
Expert (DS)        & \underline{75.8} \\
LUFFY (DS)         & 75.6 \\
GHPO (DS)          & 75.0 \\
\rowcolor{mitblue}
SMEPO (DS)          & 75.7 \\
\midrule
SFT (Gemini)       & 74.1 \\
SFT (Gemini) + RL  & 74.9 \\
Expert (Gemini)    & 75.5 \\
LUFFY (Gemini)     & 75.6 \\
GHPO (Gemini)      & 75.3 \\
\rowcolor{mitblue}
SMEPO (Gemini)      & \textbf{75.9} \\
\bottomrule
\end{tabular}
\end{subtable}
\hfill
\begin{minipage}[t]{0.60\textwidth}
\centering

\begin{subtable}[t]{0.48\textwidth}
\centering
\caption*{\textbf{Qwen3-8B-Base Code}}
\begin{tabular}{lc}
\toprule
Method & MMLU \\
\midrule
Qwen3-8B-Base & \underline{75.7} \\
GRPO          & \underline{75.7} \\
SFT           & 74.4 \\
Expert        & \textbf{75.8} \\
\rowcolor{mitblue}
SMEPO          & \textbf{75.8} \\
\bottomrule
\end{tabular}
\end{subtable}
\hfill
\begin{subtable}[t]{0.48\textwidth}
\centering
\caption*{\textbf{Qwen2.5-7B Math}}
\begin{tabular}{lc}
\toprule
Method & MMLU \\
\midrule
Qwen2.5-7B  & 72.0 \\
GRPO        & 72.2 \\
SFT    & 71.8 \\
Expert & \underline{72.4} \\
\rowcolor{mitblue}
SMEPO   & \textbf{72.5} \\
\bottomrule
\end{tabular}
\end{subtable}

\vspace{1.0em}

\begin{subtable}[t]{0.48\textwidth}
\centering
\caption*{\textbf{Qwen3-8B-Base Agent}}
\begin{tabular}{lc}
\toprule
Method & MMLU \\
\midrule
Qwen3-8B-Base & \underline{75.7} \\
GRPO          & \textbf{75.8} \\
SFT           & 75.0 \\
Expert        & 75.6 \\
\rowcolor{mitblue}
SMEPO          & \textbf{75.8} \\
\bottomrule
\end{tabular}
\end{subtable}
\hfill
\begin{subtable}[t]{0.48\textwidth}
\centering
\caption*{\textbf{R1-Distill-Qwen-7B Math}}
\begin{tabular}{lc}
\toprule
Method & MMLU \\
\midrule
R1-Distill-7B & 50.7 \\
GRPO          & 50.6 \\
Expert        & \underline{50.8} \\
\rowcolor{mitblue}
SMEPO          & \textbf{51.3} \\
\bottomrule
\end{tabular}
\end{subtable}

\end{minipage}

\caption{\textbf{General-domain sanity check.}
MMLU-Overall results for the math, code, and agent runs. SMEPO preserves broad general-domain performance across settings and often stays flat or slightly improves, while SFT tends to cause a clearer drop. Best results are in bold; second-best results are underlined.}
\label{tab:supp-mmlu}
\end{table*}
\begin{table*}[t]
\centering
\small
\setlength{\tabcolsep}{5.5pt}
\begin{tabular}{lllccc}
\toprule
Model & Task & Expert trace & GRPO Avg. & Expert Avg. & $\Delta$ \\
\midrule
Qwen3-8B-Base              & Math  & DeepSeek-R1 & 47.6 & 41.0 & -6.6 \\
Qwen3-8B-Base              & Math  & Gemini-3-Pro      & 47.6 & 40.8 & -6.8 \\
Qwen2.5-7B                 & Math  & DeepSeek-R1 & 42.4 & 38.6 & -3.8 \\
R1-Distill-Qwen-7B         & Math  & DeepSeek-R1 & 61.8 & 60.4 & -1.4 \\
Qwen3-8B-Base              & Code  & DeepSeek-R1 & 56.5 & 55.7 & -0.8 \\
Qwen3-8B-Base              & Agent & DeepSeek-R1 & 45.6 & 23.2 & -22.4 \\
\bottomrule
\end{tabular}
\caption{\textbf{Diagnostic summary of full-expert-trace conditioning.}
All values are taken from Table~\ref{tab:main-results}. Avg. denotes the task-level average used in the main table: six math benchmarks for Math, three code benchmarks for Code, and Bamboogle accuracy for Agent. $\Delta$ is Expert Avg. minus GRPO Avg.; negative values indicate that full expert-trace conditioning underperforms vanilla GRPO.}
\label{tab:full_trace_diagnosis}
\end{table*}

\section{Additional Analysis}
\label{app:additional-analysis}

The main text reports ablation results comparing different ways of exposing or masking expert traces. In this section, we provide additional analyses to better understand the behavior behind these ablations.

\subsection{Code-copy diagnostics.}
\label{app:code-copy}
For code tasks, the reward-relevant content is usually a complete executable solution rather than a short numeric answer or entity. We therefore measure whether expert guidance encourages direct copying using two complementary metrics.

Let $G_{\mathrm{rollout}}$ denote the model rollout and $G_{\mathrm{code}}$ denote the generated code block, $T_{\mathrm{vis}}$ denote the portion of the expert trace that is visible in the prompt, and $T_{\mathrm{code}}$ denote the original expert code block. We define \emph{Visible-trace overlap} as
\[
    \mathrm{VisibleTraceOverlap}
    =
    \frac{
    \mathrm{LCCS}\!\left(G_{\mathrm{rollout}}, T_{\mathrm{vis}}\right)
    }{
    |G_{\mathrm{rollout}}|
    },
\]
where $\mathrm{LCCS}(\cdot,\cdot)$ is the length of the longest common contiguous subsequence and $|G_{\mathrm{rollout}}|$ is the rollout length. This metric captures whether the model copies content that was actually exposed in the prompt.

We further define \emph{Expert-code similarity} as
\[
    \mathrm{ExpertCodeSimilarity}
    =
    \frac{
    \mathrm{LCCS}\!\left(G_{\mathrm{code}}, T_{\mathrm{code}}\right)
    }{
    |G_{\mathrm{code}}|
    }.
\]
This metric measures similarity between the generated code block and the original expert code block, independent of whether that code block was fully visible or masked in the prompt.

\begin{figure}[t]
    \centering
    \includegraphics[width=0.72\linewidth]{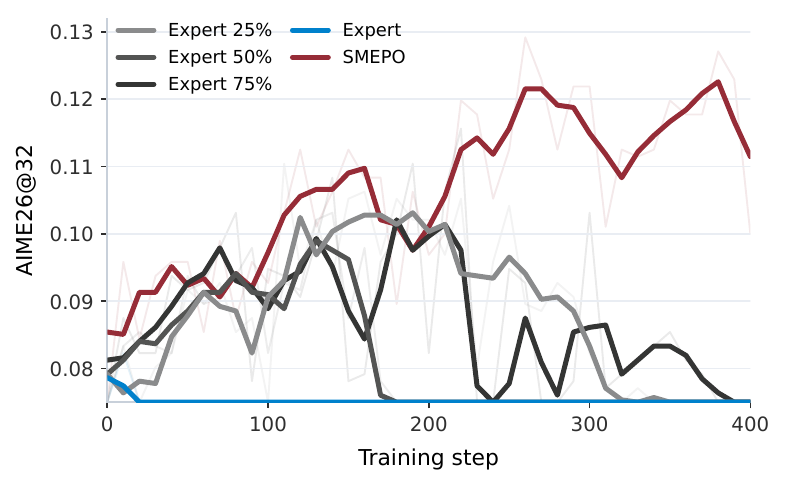}
    \caption{\textbf{Effect of partial expert-trace exposure on AIME26.}
    Coarse prefix exposure reduces the harm of full-expert-trace conditioning, but its effect is non-monotonic and remains below SMEPO, suggesting that the key issue is not trace length alone.}
    \label{fig:partial-exposure-curve}
\end{figure}

\subsection{Partial exposure.}
Figure~\ref{fig:partial-exposure-curve} left shows that exposing only a prefix of the expert trace is a useful but incomplete control. Prefix exposure reduces the amount of expert content visible to the policy, and therefore can reduce the harm of full-expert-trace conditioning. However, it does not explicitly distinguish guidance from reward-relevant content. Since reward-relevant spans can appear at different positions across traces, a fixed prefix ratio may remove useful later reasoning steps while still leaving exploitable intermediate values or answer-related content visible. This explains the non-monotonic behavior of prefix exposure: shorter traces are not necessarily safer, and longer traces are not necessarily more useful. SMEPO instead keeps the trace organization whenever possible and masks spans selected by reward-relevant rules, which gives a more direct control over the harmful channel.

\subsection{Why masking numbers outperforms answer-only masking.}
The ablation results in the main text show that masking only the final answer is not sufficient. In math traces, the final answer is often only one instance of a broader value channel. Expert solutions may repeat the answer value, introduce intermediate quantities that nearly determine the answer, or include arithmetic correlations that make the hidden answer easy to recover. As a result, answer-only masking can still leave enough numeric evidence for the policy to exploit the trace without learning the underlying reasoning process. SMEPO masks numeric content throughout the expert trace, removing this broader reward-relevant payload while preserving equation templates, decomposition patterns, and surrounding explanatory text. This is why numeric masking outperforms answer-only masking: it targets the distributed value information that makes expert traces too easy to exploit.

\section{Method Overhead}
\label{app:method-overhead}

We measure the offline preprocessing overhead of SMEPO on the same CPU node for all domains. The machine has dual Intel Xeon Gold 6442Y processors, with 48 physical cores and 96 logical CPUs. All measurements are performed without GPU acceleration. We use \texttt{/usr/bin/time -v} to measure the end-to-end wall-clock time of each preprocessing script, including reading the original data, applying the corresponding masking rule, and writing the processed output files.

For math, the preprocessing script masks numeric reward-relevant spans in expert reasoning traces. For code, the script masks executable code blocks in expert solutions while preserving the surrounding natural-language guidance. For agentic search, the script masks answer-relevant entities in teacher search rounds using rule-based answer/entity extraction and spaCy NER. These preprocessing steps are performed once offline before RL training, and the resulting masked data are reused throughout training.

Table~\ref{tab:method-overhead} shows that SMEPO adds only a small offline preprocessing cost. Processing 1,024 examples takes 5.96 seconds for math, 1.80 seconds for code, and 29.42 seconds for agentic search. The search setting is slower because it performs entity-level masking with spaCy NER~\citep{spacy}, whereas math and code use lightweight rule-based masking. Since masking is applied once before training, this cost is amortized over all RL updates and is negligible compared with rollout generation and model optimization.

\section{Case Studies}
\label{app:case-studies}

Because the original model outputs are in Markdown, we present the case studies in Markdown format without converting them into LaTeX or otherwise altering their content.

\subsection{Annotated Case Summaries}
\label{app:case-studies-summaries}

We first present annotated summaries of three representative case studies. These summaries are drawn from the full rollouts in Appendix~\ref{app:case-studies-full} and highlight the most diagnostic parts of each example: what the unmasked expert hint exposes, how the Expert rollout responds, where the vanilla GRPO rollout fails, and how SMEPO uses the masked trace.

The summaries are not separate examples from the full cases below. Instead, they provide a compact, analysis-oriented view of the same rollouts before the complete transcripts are shown.
\begin{table}[t]
\centering
\small
\setlength{\tabcolsep}{5.5pt}
\begin{tabular}{lrrrr}
\toprule
Domain & Examples & Wall-clock time & Time / example & Peak memory (MiB) \\
\midrule
Math & 1,024 & 5.96s & 5.82ms & 365 \\
Code & 1,024 & 1.80s & 1.76ms & 727 \\
Agentic search & 1,024 & 29.42s & 28.7ms & 850 \\
\bottomrule
\end{tabular}
\vspace{0.5em}
\caption{\textbf{Offline preprocessing overhead.}
We measure end-to-end preprocessing time with \texttt{/usr/bin/time -v} on the same CPU node for all domains. Timing includes script startup, reading the original data, applying SMEPO masking, validation checks when present, and writing the processed JSONL/parquet outputs.}
\label{tab:method-overhead}
\end{table}


\tcbset{
    SMEPOcasebase/.style={
        enhanced,
        boxrule=0.55pt,
        arc=2pt,
        left=5pt,
        right=5pt,
        top=4pt,
        bottom=4pt,
        before skip=0.35em,
        after skip=0.35em,
        fonttitle=\bfseries,
        coltitle=black,
        breakable=true,
    },
    SMEPOproblem/.style={
        SMEPOcasebase,
        colback=gray!5,
        colframe=gray!55,
        colbacktitle=gray!25,
    },
    SMEPOhint/.style={
        SMEPOcasebase,
        colback=orange!5,
        colframe=orange!65!black,
        colbacktitle=orange!20,
    },
    SMEPOexpert/.style={
        SMEPOcasebase,
        colback=red!4,
        colframe=red!55,
        colbacktitle=red!25,
    },
    SMEPOrl/.style={
        SMEPOcasebase,
        colback=gray!5,
        colframe=gray!55,
        colbacktitle=gray!25,
    },
    SMEPOmethod/.style={
        SMEPOcasebase,
        colback=blue!4,
        colframe=blue!55,
        colbacktitle=blue!18,
    },
    SMEPOanalysis/.style={
        SMEPOcasebase,
        colback=yellow!4,
        colframe=yellow!45!black,
        colbacktitle=yellow!20,
    }
}

\lstdefinestyle{SMEPOcode}{
    basicstyle=\ttfamily\footnotesize,
    breaklines=true,
    breakatwhitespace=false,
    columns=fullflexible,
    keepspaces=true,
    showstringspaces=false,
    frame=none,
    aboveskip=0pt,
    belowskip=0pt,
    xleftmargin=0pt,
    xrightmargin=0pt
}

\lstdefinestyle{SMEPOtrace}{
    basicstyle=\ttfamily\footnotesize,
    breaklines=true,
    breakatwhitespace=true,
    columns=fullflexible,
    keepspaces=true,
    showstringspaces=false,
    frame=none,
    aboveskip=0pt,
    belowskip=0pt,
    xleftmargin=0pt,
    xrightmargin=0pt
}

\subsubsection*{Annotated Summary: Math Reasoning with Expert Traces}
\label{app:case-study-math}

\begin{tcolorbox}[SMEPOproblem,title={Problem}]
Given \(f(f(n))=2n\) and \(f(4n+1)=4n+3\), find \(f(1000)\).

Ground truth: \(\boxed{1016}\).
\end{tcolorbox}

\begin{tcolorbox}[SMEPOhint,title={Unmasked Expert Hint}]
The unmasked expert hint exposes the decisive numeric path and the final answer-relevant computation:
\begin{lstlisting}[style=SMEPOtrace]
Decompose 1000 = 2^3 * 125.
125 is odd and of the form 4k + 1
(125 = 4 * 31 + 1), so f(125) = 4 * 31 + 3 = 127.
Therefore, f(1000) = 2^3 * f(125) = 8 * 127 = 1016.
\end{lstlisting}
Thus the hint already contains the target decomposition, the intermediate value \(f(125)=127\), and the final answer \(1016\).
\end{tcolorbox}

\begin{tcolorbox}[SMEPOexpert,title={Expert Output}]
After conditioning on the unmasked expert trace, the rollout does not show an independent derivation. It simply emits the final boxed answer:
\[
\boxed{1016}
\]
\end{tcolorbox}

\begin{tcolorbox}[SMEPOrl,title={GRPO Output}]
Without expert guidance, the rollout does not recover the useful decomposition rule. It repeatedly tries to solve the problem through a preimage search:
\begin{lstlisting}[style=SMEPOtrace]
Let's denote f(1000) = k. Then, using the first property, we have f(k) = 2000.

We need to find k such that f(k) = 2000. Since 2000 is even, we need to find an odd number m such that f(m) = 2001. Using the second property, we have f(4k + 1) = 4k + 3. So, we need to find k such that 4k + 3 = 2001.

Solving for k:
4k + 3 = 2001
4k = 1998
k = 499.5

Since k must be an integer, we need to adjust our approach.
\end{lstlisting}
The rollout repeats this failed pattern, repeatedly forcing even targets such as \(2000\) into a rule that applies to inputs of the form \(4n+1\).
\end{tcolorbox}

\begin{tcolorbox}[SMEPOmethod,title={SMEPO Output}]
The masked trace removes the concrete numbers but preserves the decomposition template. The SMEPO rollout reconstructs the missing numeric content and derives the scaling rule it uses:
\begin{lstlisting}[style=SMEPOtrace]
1000 is an even number, and we can express it as
1000 = 2^3 * 125. Here, 125 is odd and can be written as
125 = 4 * 31 + 1.

Since 125 = 4 * 31 + 1, we have
f(125) = 4 * 31 + 3 = 127.

First, derive the scaling property. From f(f(n)) = 2n,
applying the same identity to f(n) gives
f(f(f(n))) = 2f(n). Since f(f(n)) = 2n, the left-hand side
is also f(2n). Hence f(2n) = 2f(n), and by induction
f(2^k * m) = 2^k * f(m).

Since 1000 = 2^3 * 125, we get
f(1000) = f(2^3 * 125) = 2^3 * f(125)
        = 8 * 127 = 1016.
\end{lstlisting}
\[
\boxed{1016}
\]
\end{tcolorbox}

\begin{tcolorbox}[SMEPOanalysis,title={Analysis}]
This case shows why the difference between the hint and the rollout matters. The unmasked expert hint contains the answer-relevant computation, including \(1000=2^3\cdot125\), \(125=4\cdot31+1\), \(f(125)=127\), and \(8\cdot127=1016\). Once this information is visible, the resulting Expert rollout can obtain reward by directly emitting \(\boxed{1016}\). The rollout is correct, but it does not demonstrate that the model has learned or derived a reusable reasoning procedure.

The vanilla GRPO rollout fails for a specific reason: it tries to invert \(f\) by setting \(f(1000)=k\) and then matching \(f(k)=2000\) to the rule \(f(4n+1)=4n+3\). This repeatedly produces nonintegral or mismatched congruence conditions, such as \(4k+3=2001\). The missing insight is not just the final numeric answer, but the structural reduction: first derive \(f(2n)=2f(n)\) from \(f(f(n))=2n\), then use induction to obtain \(f(2^k m)=2^k f(m)\), and finally reduce \(1000\) to its odd component \(125\).

SMEPO keeps this useful structural plan while masking the concrete answer-relevant values. The rollout still has to recover \(1000=2^3\cdot125\), identify \(125=4\cdot31+1\), compute \(f(125)=127\), derive the scaling rule \(f(2^k m)=2^k f(m)\), and combine these steps to obtain \(1016\). Thus the masked trace provides the reasoning route without directly exposing the reward-relevant numeric path as an answer channel.
\end{tcolorbox}

\subsubsection*{Annotated Summary: Code Synthesis with Expert Traces}
\label{app:case-study-code}

\begin{tcolorbox}[SMEPOproblem,title={Problem}]
Given \(N\) points in a plane, choose a nonzero direction \((p,q)\). Points can be collected at zero additional cost when the previously collected point is exactly one step \((p,q)\) behind. The goal is to choose \((p,q)\) to minimize the number of chains needed to collect all points.
\end{tcolorbox}

\begin{tcolorbox}[SMEPOhint,title={Unmasked Expert Hint}]
The unmasked expert hint exposes a complete executable solution. Its core idea is to enumerate all offsets induced by point pairs and count how many chains start under each offset:
\begin{lstlisting}[style=SMEPOcode]
points_set = set(points)
diff_set = set()

for i in range(n):
    x_i, y_i = points[i]
    for j in range(n):
        if i != j:
            x_j, y_j = points[j]
            dx = x_j - x_i
            dy = y_j - y_i
            diff_set.add((dx, dy))

min_chains = None

for dx, dy in diff_set:
    current = 0
    for (x, y) in points:
        if (x - dx, y - dy) not in points_set:
            current += 1
    if min_chains is None or current < min_chains:
        min_chains = current

print(min_chains)
\end{lstlisting}
\end{tcolorbox}

\begin{tcolorbox}[SMEPOexpert,title={Expert Output}]
After conditioning on the unmasked expert trace, the rollout directly emits the solution implementation. The output is essentially the expert algorithm:
\begin{lstlisting}[style=SMEPOcode]
n = int(input())
points = [tuple(map(int, input().split())) for _ in range(n)]

if n == 1:
    print(1)
    exit()

points_set = set(points)
diff_set = set()

for i in range(n):
    x_i, y_i = points[i]
    for j in range(n):
        if i != j:
            x_j, y_j = points[j]
            dx = x_j - x_i
            dy = y_j - y_i
            diff_set.add((dx, dy))

min_chains = None

for dx, dy in diff_set:
    current = 0
    for (x, y) in points:
        if (x - dx, y - dy) not in points_set:
            current += 1
    if min_chains is None or current < min_chains:
        min_chains = current

print(min_chains)
\end{lstlisting}
\end{tcolorbox}

\begin{tcolorbox}[SMEPOrl,title={GRPO Output}]
The vanilla GRPO rollout instead produces an invalid bounded search over small \((p,q)\) values:
\begin{lstlisting}[style=SMEPOcode]
min_cost = float('inf')
for p in range(-10, 11):
    for q in range(-10, 11):
        if p == 0 and q == 0:
            continue
        cost = 0
        collected = set()
        for i in range(N):
            for j in range(i + 1, N):
                if (balls[i][0], balls[i][1]) in collected and \
                   (balls[j][0], balls[j][1]) in collected:
                    continue
                if (balls[i][0] - balls[j][0]) != \
                   p * (balls[i][1] - balls[j][1]) // q or \
                   (balls[i][1] - balls[j][1]) % q != 0:
                    cost += 1
                    collected.add((balls[i][0], balls[i][1]))
                    collected.add((balls[j][0], balls[j][1]))
            min_cost = min(min_cost, cost)
print(min_cost)
\end{lstlisting}
This code is not equivalent to the chain-cover objective.
\end{tcolorbox}

\begin{tcolorbox}[SMEPOmethod,title={SMEPO Output}]
The masked trace hides the full code block but keeps the algorithmic procedure. The SMEPO rollout reconstructs the key logic: enumerate point-pair offsets, then count points without predecessors.
\begin{lstlisting}[style=SMEPOcode]
points_set = set(tuple(point) for point in points)

if n == 1:
    print(1)
    return

min_cost = float('inf')
dx_dy_pairs = set()

for i in range(n):
    for j in range(i + 1, n):
        dx = points[j][0] - points[i][0]
        dy = points[j][1] - points[i][1]
        dx_dy_pairs.add((dx, dy))

for dx, dy in dx_dy_pairs:
    chains = 0
    for x, y in points:
        if (x - dx, y - dy) not in points_set:
            chains += 1
    min_cost = min(min_cost, chains)

print(min_cost)
\end{lstlisting}
\end{tcolorbox}

\begin{tcolorbox}[SMEPOanalysis,title={Analysis}]
This case highlights a code-specific form of reward-relevant content. In code synthesis, the unmasked expert hint does not merely reveal an intermediate clue; it contains a complete executable implementation. The resulting Expert output directly emits the same algorithm. This is correct, but the reward can be achieved by reproducing the exposed code body rather than by learning the underlying programming strategy.

The vanilla GRPO output illustrates a different failure mode. It treats the direction \((p,q)\) as something that can be brute-forced in the bounded range \([-10,10]^2\), although the coordinates can be as large as \(10^9\) and the optimal direction can be any offset between two input points. It also uses a slope-like integer-division test involving \(q\), which is not the same as checking whether one point is exactly one step \((p,q)\) after another. This can fail for horizontal directions when \(q=0\) and is conceptually wrong even when \(q\neq0\). Finally, it increments cost inside a pairwise loop with a ``collected'' set, whereas the correct objective for a fixed offset is to count chain starts: points whose predecessor \((x-p,y-q)\) is absent.

SMEPO masks the directly copyable code body while preserving the procedural outline. The generated solution is not necessarily textually identical to the expert, but it recovers the essential invariant: for each candidate offset, every chain contributes exactly one start point. This is the intended effect of masking in code: remove the executable answer while retaining enough guidance for the model to synthesize the algorithm.
\end{tcolorbox}

\subsubsection*{Annotated Summary: Agentic Search with Expert Traces}
\label{app:case-study-agent}

\begin{tcolorbox}[SMEPOproblem,title={Problem}]
What age was the American social entrepreneur who was one of eight killed in the 2015 Amtrak train derailment from Washington, D.C. to New York City?

Ground truth: \(\boxed{39}\).
\end{tcolorbox}

\begin{tcolorbox}[SMEPOhint,title={Unmasked Expert Hint}]
The unmasked expert hint reveals both the target entity and the answer:
\begin{lstlisting}[style=SMEPOtrace]
Round 2:
From the list of victims, Rachel Jacobs was identified as the CEO of a technology education company, which aligns with the description of a social entrepreneur. Next, I'll search for her age at the time of the incident.

<search> Find the age of Rachel Jacobs at the time of the 2015 Amtrak train derailment. </search>

Round 3:
Based on the information, Rachel Jacobs was 39 years old at the time of the 2015 Amtrak train derailment. She was the American social entrepreneur among the victims.

<answer>39</answer>
\end{lstlisting}
\end{tcolorbox}

\begin{tcolorbox}[SMEPOexpert,title={Expert Output}]
After conditioning on the unmasked expert trace, the rollout directly emits the answer:
\[
\texttt{Answer: <answer>39</answer>}
\]
\end{tcolorbox}

\begin{tcolorbox}[SMEPOrl,title={GRPO Output}]
The vanilla GRPO rollout identifies the derailment but repeatedly searches for generic victim lists. It fails to pivot to the discriminative entity description:
\begin{lstlisting}[style=SMEPOtrace]
<search> Amtrak derailment 2015 </search>

From Doc 1, I see that the Amtrak derailment in 2015 involved a train traveling from Washington, D.C. to New York City and resulted in 8 deaths and over 200 injuries. However, I don't see the names of the individuals who were killed.

<search> Amtrak derailment 2015 victims </search>

From Doc 1, I see that 8 people were killed in the derailment, but I still don't have the names of the individuals. I will need to search for a more detailed list of the victims.

<search> names of the 8 people killed in the 2015 Amtrak derailment </search>

From Doc 1, I still don't have the names of the individuals. I will need to search for a more detailed list of the victims.

<search> 2015 Amtrak derailment victim list </search>

From Doc 1, I still don't have the names of the individuals. I will need to search for a more detailed list of the victims.

<search> 2015 Philadelphia train derailment victim names </search>
\end{lstlisting}
\end{tcolorbox}

\begin{tcolorbox}[SMEPOmethod,title={SMEPO Output}]
The masked trace removes answer-relevant entities such as the target name and age, but preserves the search plan. The SMEPO rollout reformulates the second query around the entity description:
\begin{lstlisting}[style=SMEPOtrace]
<search> Find information about the 2015 Amtrak train derailment from Washington, D.C. to New York City, including the list of victims killed in the incident. </search>

From the information, I've found that the Amtrak train derailment occurred on May 12, 2015, and 8 people were killed, including a social entrepreneur. Next, I'll search for the name of the social entrepreneur and her age at the time of the incident.

<search> Find the name of the American social entrepreneur who was one of the eight killed in the 2015 Amtrak train derailment from Washington, D.C. to New York City, and her age at the time of the incident. </search>
\end{lstlisting}
The retrieved evidence identifies Rachel Jacobs and states that she was killed at age \(39\). The rollout returns:
\[
\texttt{<answer>39</answer>}.
\]
\end{tcolorbox}

\begin{tcolorbox}[SMEPOanalysis,title={Analysis}]
This case cleanly separates answer exposure from useful search procedure. The unmasked expert hint directly reveals the target entity, Rachel Jacobs, and the final age, \(39\). The resulting Expert rollout is therefore extremely short: it simply emits \(\texttt{<answer>39</answer>}\). This is precisely the kind of reward-relevant channel that can make full expert traces problematic. The policy can achieve reward by finalizing exposed information rather than by performing the search.

The vanilla GRPO rollout finds the correct event but fails to use the key constraint in the question. It repeatedly searches for generic victim lists, such as ``Amtrak derailment 2015 victims'' and ``names of the 8 people killed.'' These searches return descriptions of the derailment but do not surface the needed entity. The rollout loops over near-duplicate search intents instead of pivoting to the more specific clue: the victim was an American social entrepreneur.

SMEPO removes the target entity and final answer from the trace, but preserves the useful action sequence. After identifying the derailment, the model learns to reformulate the query around the discriminative entity description and the desired attribute: the name of the American social entrepreneur and her age. This query retrieves the target entity and supports the final answer \(39\). Thus the masked trace guides the search strategy without directly revealing the reward-relevant entity.
\end{tcolorbox}

\subsection{Full Case Study Rollouts}

We then include the full rollout versions of the same case studies for completeness and transparency. The annotated summaries above quote only the most relevant spans, while the full rollouts preserve the surrounding context, including the original hints and model outputs.



\tcbset{
    SMEPOfullbase/.style={
        enhanced,
        breakable,
        boxrule=0.55pt,
        arc=2pt,
        left=6pt,
        right=6pt,
        top=5pt,
        bottom=5pt,
        before skip=0.4em,
        after skip=0.4em,
        fonttitle=\bfseries,
        coltitle=black,
    },
    SMEPOfullproblem/.style={
        SMEPOfullbase,
        colback=gray!5,
        colframe=gray!55,
        colbacktitle=gray!25,
    },
    SMEPOfullexpert/.style={
        SMEPOfullbase,
        colback=red!4,
        colframe=red!55,
        colbacktitle=red!20,
    },
    SMEPOfullrl/.style={
        SMEPOfullbase,
        colback=gray!5,
        colframe=gray!55,
        colbacktitle=gray!25,
    },
    SMEPOfullSMEPO/.style={
        SMEPOfullbase,
        colback=blue!4,
        colframe=blue!55,
        colbacktitle=blue!15,
    },
    SMEPOfulloutcome/.style={
        SMEPOfullbase,
        colback=gray!5,
        colframe=gray!55,
        colbacktitle=gray!25,
    }
}

\lstdefinestyle{SMEPOfullcode}{
    basicstyle=\ttfamily\footnotesize,
    breaklines=true,
    breakatwhitespace=false,
    columns=fullflexible,
    keepspaces=true,
    showstringspaces=false,
    frame=none,
    aboveskip=2pt,
    belowskip=2pt,
    xleftmargin=0pt,
    xrightmargin=0pt
}

\lstdefinestyle{SMEPOfulltrace}{
    basicstyle=\ttfamily\footnotesize,
    breaklines=true,
    breakatwhitespace=true,
    columns=fullflexible,
    keepspaces=true,
    showstringspaces=false,
    frame=none,
    aboveskip=2pt,
    belowskip=2pt,
    xleftmargin=0pt,
    xrightmargin=0pt
}

\label{app:case-studies-full}

\subsubsection*{Full Rollout:  Functional Equation Reasoning}

\begin{tcolorbox}[SMEPOfullproblem,title={Question}]
\begin{lstlisting}[style=SMEPOfulltrace]
The function $f(n)$ is defined on the positive integers such that $f(f(n)) = 2n$ and $f(4n + 1) = 4n + 3$ for all positive integers $n.$  Find $f(1000).$
\end{lstlisting}
\end{tcolorbox}

\begin{tcolorbox}[SMEPOfullproblem,title={Ground Truth}]
\begin{lstlisting}[style=SMEPOfulltrace]
1016
\end{lstlisting}
\end{tcolorbox}

\begin{tcolorbox}[SMEPOfullexpert,title={Unmasked Expert Hint}]
\begin{lstlisting}[style=SMEPOfulltrace]
To solve for \( f(1000) \), we analyze the function \( f \) defined by the conditions \( f(f(n)) = 2n \) and \( f(4n + 1) = 4n + 3 \). 

1. **Odd Numbers**:
   - For \( n = 4k + 1 \), \( f(n) = 4k + 3 \).
   - For \( n = 4k + 3 \), \( f(n) = 8k + 2 \).

2. **Even Numbers**:
   - If \( n \) is even, express \( n \) as \( 2^k \cdot m \) where \( m \) is odd. Then \( f(n) = 2^k \cdot f(m) \).

**Application to \( n = 1000 \)**:
- Decompose \( 1000 = 2^3 \cdot 125 \).
- \( 125 \) is odd and of the form \( 4k + 1 \) (\( 125 = 4 \cdot 31 + 1 \)), so \( f(125) = 4 \cdot 31 + 3 = 127 \).
- Therefore, \( f(1000) = 2^3 \cdot f(125) = 8 \cdot 127 = 1016 \).

Verification:
- \( f(1016) = 2^3 \cdot f(127) \). Since \( 127 = 4 \cdot 31 + 3 \), \( f(127) = 8 \cdot 31 + 2 = 250 \).
- Thus, \( f(1016) = 8 \cdot 250 = 2000 \), confirming \( f(f(1000)) = 2 \cdot 1000 \).
\end{lstlisting}
\end{tcolorbox}

\begin{tcolorbox}[SMEPOfullSMEPO,title={Masked Expert Hint}]
\begin{lstlisting}[style=SMEPOfulltrace]
To solve for \( f([NUMBER]) \), we analyze the function \( f \) defined by the conditions \( f(f(n)) = [NUMBER]n \) and \( f([NUMBER]n + [NUMBER]) = [NUMBER]n + [NUMBER] \). 

**Odd Numbers**:
   - For \( n = [NUMBER]k + [NUMBER] \), \( f(n) = [NUMBER]k + [NUMBER] \).
   - For \( n = [NUMBER]k + [NUMBER] \), \( f(n) = [NUMBER]k + [NUMBER] \).

**Even Numbers**:
   - If \( n \) is even, express \( n \) as \( [NUMBER]^k \cdot m \) where \( m \) is odd. Then \( f(n) = [NUMBER]^k \cdot f(m) \).

**Application to \( n = [NUMBER] \)**:
- Decompose \( [NUMBER] = [NUMBER]^[NUMBER] \cdot [NUMBER] \).
- \( [NUMBER] \) is odd and of the form \( [NUMBER]k + [NUMBER] \) (\( [NUMBER] = [NUMBER] \cdot [NUMBER] + [NUMBER] \)), so \( f([NUMBER]) = [NUMBER] \cdot [NUMBER] + [NUMBER] = [NUMBER] \).
- Therefore, \( f([NUMBER]) = [NUMBER]^[NUMBER] \cdot f([NUMBER]) = [NUMBER] \cdot [NUMBER] = [NUMBER] \).

Verification:
- \( f([NUMBER]) = [NUMBER]^[NUMBER] \cdot f([NUMBER]) \). Since \( [NUMBER] = [NUMBER] \cdot [NUMBER] + [NUMBER] \), \( f([NUMBER]) = [NUMBER] \cdot [NUMBER] + [NUMBER] = [NUMBER] \).
- Thus, \( f([NUMBER]) = [NUMBER] \cdot [NUMBER] = [NUMBER] \), confirming \( f(f([NUMBER])) = [NUMBER] \cdot [NUMBER] \).
\end{lstlisting}
\end{tcolorbox}

\begin{tcolorbox}[SMEPOfullexpert,title={Expert Output}]
\begin{lstlisting}[style=SMEPOfulltrace]
\boxed{1016}
\end{lstlisting}
\end{tcolorbox}

\begin{tcolorbox}[SMEPOfullrl,title={GRPO Output}]
\begin{lstlisting}[style=SMEPOfulltrace]
To solve the problem, we need to determine the value of \( f(1000) \) given the properties of the function \( f \):
1. \( f(f(n)) = 2n \)
2. \( f(4n + 1) = 4n + 3 \)

Let's proceed step by step.

### Step 1: Understand the function properties
- The first property \( f(f(n)) = 2n \) tells us that applying the function twice doubles the input.
- The second property \( f(4n + 1) = 4n + 3 \) gives us the value of the function for numbers of the form \( 4n + 1 \).

### Step 2: Find a pattern or a way to express \( f(n) \) for other values
We need to find \( f(1000) \). Notice that 1000 is not of the form \( 4n + 1 \), so we need to use the first property to find a way to express \( f(1000) \).

Let's denote \( f(1000) = k \). Then, using the first property, we have \( f(k) = 2000 \).

Now, we need to find \( k \). Since \( k \) is not of the form \( 4n + 1 \), we need to find a number \( m \) such that \( f(m) = 1000 \). Then, using the first property, we have \( f(f(m)) = 2m \), so \( f(1000) = 2m \).

To find \( m \), we need to express 1000 in a form that we can use the second property. Notice that 1000 is even, so it is not directly useful with the second property. However, we can use the first property to find a relationship.

Since \( f(f(n)) = 2n \), we can try to find a number \( n \) such that \( f(n) = 1000 \). Let's denote \( f(n) = 1000 \). Then, using the first property, we have \( f(1000) = 2n \).

We need to find \( n \) such that \( f(n) = 1000 \). Since 1000 is even, we need to find an odd number \( m \) such that \( f(m) = 1001 \) (the closest odd number to 1000). Using the second property, we have \( f(4k + 1) = 4k + 3 \). So, we need to find \( k \) such that \( 4k + 3 = 1001 \).

Solving for \( k \):
\[ 4k + 3 = 1001 \]
\[ 4k = 998 \]
\[ k = 249.5 \]

Since \( k \) must be an integer, we need to adjust our approach. We can use the fact that \( f(f(n)) = 2n \) to find a direct relationship. Let's denote \( f(1000) = k \). Then, \( f(k) = 2000 \).

We need to find \( k \) such that \( f(k) = 2000 \). Since 2000 is even, we need to find an odd number \( m \) such that \( f(m) = 2001 \). Using the second property, we have \( f(4k + 1) = 4k + 3 \). So, we need to find \( k \) such that \( 4k + 3 = 2001 \).

Solving for \( k \):
\[ 4k + 3 = 2001 \]
\[ 4k = 1998 \]
\[ k = 499.5 \]

Since \( k \) must be an integer, we need to adjust our approach. We can use the fact that \( f(f(n)) = 2n \) to find a direct relationship. Let's denote \( f(1000) = k \). Then, \( f(k) = 2000 \).

We need to find \( k \) such that \( f(k) = 2000 \). Since 2000 is even, we need to find an odd number \( m \) such that \( f(m) = 2001 \). Using the second property, we have \( f(4k + 1) = 4k + 3 \). So, we need to find \( k \) such that \( 4k + 3 = 2001 \).

Solving for \( k \):
\[ 4k + 3 = 2001 \]
\[ 4k = 1998 \]
\[ k = 499.5 \]

...

Since \( k \) must be an integer, we need to adjust our approach. We can use the fact that \( f(f(n)) = 2n \) to find a direct relationship. Let's denote \( f(1000) = k \). Then, \( f(k) = 2000 \).

We need to find \( k \) such that \( f(k) = 2000 \). Since 2000 is even, we need to find an odd number \( m \) such that \( f(m) = 2001 \). Using the second property, we have \( f(4k + 1) = 4k + 3 \). So, we need to find \( k \) such that \( 4k + 3 = 2001 \).

Solving for \( k \):
\[ 4k + 3 = 2001 \]
\[ 4k = 1998 \]
\[ k = 499.5 \]

Since \( k \) must be an integer, we need to adjust our approach. We can use the fact that \( f(f(n)) = 2n \) to find a direct relationship. Let's denote \( f(1000) = k \). Then, \( f(k) = 2000 \).
\end{lstlisting}
\end{tcolorbox}

\begin{tcolorbox}[SMEPOfullSMEPO,title={SMEPO Output}]
\begin{lstlisting}[style=SMEPOfulltrace]
To find \( f(1000) \), we need to understand the behavior of the function \( f \) given the conditions \( f(f(n)) = 2n \) and \( f(4n + 1) = 4n + 3 \).

1. **Identify the form of \( n \):**
   - \( 1000 \) is an even number, and we can express it as \( 1000 = 2^3 \cdot 125 \). Here, \( 125 \) is odd and can be written as \( 125 = 4 \cdot 31 + 1 \).

2. **Use the given conditions:**
   - Since \( 125 = 4 \cdot 31 + 1 \), we have \( f(125) = 4 \cdot 31 + 3 = 127 \).

3. **Determine \( f(1000) \):**
   - First, derive the scaling property. From \( f(f(n)) = 2n \), applying the same identity to \( f(n) \) gives \( f(f(f(n))) = 2f(n) \). Since \( f(f(n)) = 2n \), the left-hand side is also \( f(2n) \). Hence \( f(2n) = 2f(n) \), and by induction \( f(2^k \cdot m) = 2^k \cdot f(m) \).
   - Since \( 1000 = 2^3 \cdot 125 \), we get \( f(1000) = f(2^3 \cdot 125) = 2^3 \cdot f(125) = 8 \cdot 127 = 1016 \).

Therefore, the final answer is \boxed{1016}.
\end{lstlisting}
\end{tcolorbox}

\begin{tcolorbox}[SMEPOfulloutcome,title={Outcome}]
\textbf{Expert:} Correct\quad
\textbf{GRPO:} Wrong\quad
\textbf{SMEPO:} Correct
\end{tcolorbox}

\subsubsection*{Full Rollout:  Chain Counting Over Candidate Directions}

\begin{tcolorbox}[SMEPOfullproblem,title={Question}]
\begin{lstlisting}[style=SMEPOfulltrace]
Solve the following programming problem.
Your program should read from standard input and write to standard output.

There are N balls in a two-dimensional plane. The i-th ball is at coordinates (x_i, y_i).

We will collect all of these balls, by choosing two integers p and q such that p \neq 0 or q \neq 0 and then repeating the following operation:

* Choose a ball remaining in the plane and collect it. Let (a, b) be the coordinates of this ball. If we collected a ball at coordinates (a - p, b - q) in the previous operation, the cost of this operation is 0. Otherwise, including when this is the first time to do this operation, the cost of this operation is 1.



Find the minimum total cost required to collect all the balls when we optimally choose p and q.

Constraints

* 1 \leq N \leq 50
* |x_i|, |y_i| \leq 10^9
* If i \neq j, x_i \neq x_j or y_i \neq y_j.
* All values in input are integers.

Input

Input is given from Standard Input in the following format:


N
x_1 y_1
:
x_N y_N


Output

Print the minimum total cost required to collect all the balls.

Examples

Input

2
1 1
2 2


Output

1


Input

3
1 4
4 6
7 8


Output

1


Input

4
1 1
1 2
2 1
2 2


Output

2
\end{lstlisting}
\end{tcolorbox}

\begin{tcolorbox}[SMEPOfullproblem,title={Ground Truth}]
\begin{lstlisting}[style=SMEPOfulltrace]
{"inputs": ["3\n1 4\n4 6\n7 8", "4\n1 1\n1 2\n2 1\n2 2", "2\n1 1\n2 2", "3\n1 4\n4 6\n11 8", "2\n1 2\n2 2", "4\n1 1\n1 4\n2 1\n2 2", "3\n1 4\n4 6\n11 3", "3\n0 4\n4 6\n11 3", "3\n-1 4\n4 6\n11 3", "3\n-2 4\n4 6\n11 3", "3\n-2 4\n5 6\n11 3", "3\n-2 4\n5 6\n11 4", "3\n-1 4\n5 6\n11 4", "3\n-1 4\n5 10\n11 4", "3\n-1 7\n5 10\n11 4", "3\n-1 7\n5 12\n11 4", "3\n-1 7\n5 12\n10 4", "3\n-2 7\n5 12\n10 4", "3\n-2 7\n5 16\n10 4", "3\n-2 0\n5 16\n10 4", "3\n-2 -1\n5 16\n10 4", "3\n0 0\n5 16\n10 4", "3\n-1 0\n5 16\n10 4", "3\n-1 0\n5 32\n10 4", "3\n-1 1\n5 32\n10 4", "3\n-1 2\n5 32\n10 4", "3\n-1 4\n5 32\n10 4", "3\n-1 4\n5 32\n10 2", "3\n-1 4\n5 32\n10 0", "3\n-1 4\n5 32\n20 0", "3\n-1 4\n9 32\n20 0", "3\n-1 4\n9 32\n24 0", "3\n-1 4\n9 34\n24 0", "3\n-1 6\n9 34\n24 0", "3\n-1 7\n9 34\n24 0", "3\n-2 7\n9 34\n24 0", "3\n-4 7\n9 34\n24 0", "3\n-4 7\n9 34\n24 -1", "3\n0 7\n9 34\n24 -1", "3\n0 7\n9 33\n24 -1", "3\n0 7\n9 33\n24 0", "3\n0 7\n10 33\n24 0", "3\n0 7\n10 33\n24 -1", "3\n0 7\n10 33\n1 -1", "3\n0 4\n10 33\n1 -1", "3\n0 4\n10 33\n2 -1", "3\n0 4\n10 33\n3 -1", "3\n0 4\n2 33\n3 -1", "3\n0 4\n1 33\n3 -1", "3\n0 4\n1 17\n3 -1", "3\n0 1\n1 17\n3 -1", "3\n0 1\n1 17\n6 -1", "3\n0 1\n1 17\n6 0", "3\n0 1\n0 17\n6 0", "3\n1 1\n0 17\n6 0", "3\n1 1\n0 17\n2 0", "3\n0 1\n0 17\n2 0", "3\n0 1\n0 25\n2 0", "3\n0 1\n0 25\n4 0", "3\n0 1\n0 25\n6 0", "3\n0 1\n0 32\n6 0", "3\n0 1\n1 32\n6 0", "3\n1 1\n1 32\n6 0", "3\n1 1\n1 32\n6 1", "3\n1 0\n1 32\n6 1", "3\n1 0\n1 32\n6 2", "3\n1 0\n0 32\n6 2", "3\n1 0\n0 32\n6 1", "3\n2 0\n0 32\n6 1", "3\n2 1\n0 32\n6 1", "3\n1 4\n3 6\n7 8", "2\n2 1\n2 2", "3\n1 4\n4 6\n8 8", "2\n1 2\n2 4", "3\n2 4\n4 6\n11 3", "3\n0 4\n0 6\n11 3", "3\n-1 4\n4 6\n11 2", "3\n-2 5\n4 6\n11 3", "3\n-2 4\n5 3\n11 3", "3\n-2 4\n5 10\n11 4", "3\n-1 4\n5 6\n1 4", "3\n-2 4\n5 2\n11 4", "3\n-1 7\n5 10\n21 4", "3\n-1 7\n5 12\n11 2", "3\n-1 7\n5 3\n10 4", "3\n-2 7\n6 12\n10 4", "3\n-2 7\n4 16\n10 4", "3\n-4 0\n5 16\n10 4", "3\n-2 -1\n10 16\n10 4", "3\n1 0\n5 16\n10 4", "3\n-1 0\n5 2\n10 4", "3\n-1 0\n5 32\n20 4", "3\n-1 0\n6 32\n10 4", "3\n-1 2\n5 33\n10 4", "3\n-1 4\n5 32\n10 3", "3\n-1 4\n5 32\n9 2", "3\n-1 3\n5 32\n10 0", "3\n-1 4\n5 49\n20 0", "3\n-1 4\n9 44\n20 0", "3\n-1 4\n9 19\n24 0", "3\n-1 4\n9 34\n45 0", "3\n-1 6\n9 34\n10 0", "3\n-1 7\n9 34\n24 -1"], "outputs": ["1", "2", "1", "2\n", "1\n", "3\n", "2\n", "2\n", "2\n", "2\n", "2\n", "2\n", "2\n", "2\n", "2\n", "2\n", "2\n", "2\n", "2\n", "2\n", "2\n", "2\n", "2\n", "2\n", "2\n", "2\n", "2\n", "2\n", "2\n", "2\n", "2\n", "2\n", "2\n", "2\n", "2\n", "2\n", "2\n", "2\n", "2\n", "2\n", "2\n", "2\n", "2\n", "2\n", "2\n", "2\n", "2\n", "2\n", "2\n", "2\n", "2\n", "2\n", "2\n", "2\n", "2\n", "2\n", "2\n", "2\n", "2\n", "2\n", "2\n", "2\n", "2\n", "2\n", "2\n", "2\n", "2\n", "2\n", "2\n", "2\n", "2\n", "1\n", "2\n", "1\n", "2\n", "2\n", "2\n", "2\n", "2\n", "2\n", "2\n", "2\n", "2\n", "2\n", "2\n", "2\n", "2\n", "2\n", "2\n", "2\n", "2\n", "2\n", "2\n", "2\n", "2\n", "2\n", "2\n", "2\n", "2\n", "2\n", "2\n", "2\n", "2\n"]}
\end{lstlisting}
\end{tcolorbox}

\begin{tcolorbox}[SMEPOfullexpert,title={Unmasked Expert Hint}]
\begin{lstlisting}[style=SMEPOfulltrace]
To solve this problem, we need to determine the minimum total cost required to collect all balls in a 2D plane by optimally choosing two integers p and q (not both zero). The cost is minimized by arranging the balls into the fewest possible chains, where each chain consists of consecutive points differing by (p, q).

### Approach
1. **Input Handling**: Read the number of balls and their coordinates.
2. **Special Case for Single Ball**: If there is only one ball, the cost is always 1.
3. **Generate Differences**: Compute all possible (dx, dy) pairs from the differences between every pair of distinct points.
4. **Determine Chains**: For each (dx, dy) pair, calculate the number of chains required. A chain starts at a point if the previous point (by subtracting dx and dy) is not present.
5. **Find Minimum Cost**: Iterate through all (dx, dy) pairs to find the minimum number of chains, which corresponds to the minimum cost.

### Solution Code
``\`python
n = int(input())
points = [tuple(map(int, input().split())) for _ in range(n)]

if n == 1:
    print(1)
    exit()

points_set = set(points)
diff_set = set()

for i in range(n):
    x_i, y_i = points[i]
    for j in range(n):
        if i != j:
            x_j, y_j = points[j]
            dx = x_j - x_i
            dy = y_j - y_i
            diff_set.add((dx, dy))

min_chains = None

for dx, dy in diff_set:
    current = 0
    for (x, y) in points:
        if (x - dx, y - dy) not in points_set:
            current += 1
    if min_chains is None or current < min_chains:
        min_chains = current

print(min_chains)
``\`

### Explanation
1. **Input Handling**: The code reads the number of balls and their coordinates, storing them in a list and a set for quick lookups.
2. **Special Case Handling**: If there's only one ball, the cost is directly output as 1.
3. **Generate Differences**: For each pair of distinct points, the differences (dx, dy) are computed and stored in a set to avoid duplicates.
4. **Chain Calculation**: For each (dx, dy) pair, the code checks each point to see if it can start a new chain by verifying if the previous point (x-dx, y-dy) exists in the set of points. The count of such starting points gives the number of chains.
5. **Find Minimum**: The minimum number of chains across all (dx, dy) pairs is determined and printed as the result.
\end{lstlisting}
\end{tcolorbox}

\begin{tcolorbox}[SMEPOfullSMEPO,title={Masked Expert Hint}]
\begin{lstlisting}[style=SMEPOfulltrace]
To solve this problem, we need to determine the minimum total cost required to collect all balls in a 2D plane by optimally choosing two integers p and q (not both zero). The cost is minimized by arranging the balls into the fewest possible chains, where each chain consists of consecutive points differing by (p, q).

### Approach
1. **Input Handling**: Read the number of balls and their coordinates.
2. **Special Case for Single Ball**: If there is only one ball, the cost is always 1.
3. **Generate Differences**: Compute all possible (dx, dy) pairs from the differences between every pair of distinct points.
4. **Determine Chains**: For each (dx, dy) pair, calculate the number of chains required. A chain starts at a point if the previous point (by subtracting dx and dy) is not present.
5. **Find Minimum Cost**: Iterate through all (dx, dy) pairs to find the minimum number of chains, which corresponds to the minimum cost.

### Solution Code
``\`python
[CODE]
``\`

### Explanation
1. **Input Handling**: The code reads the number of balls and their coordinates, storing them in a list and a set for quick lookups.
2. **Special Case Handling**: If there's only one ball, the cost is directly output as 1.
3. **Generate Differences**: For each pair of distinct points, the differences (dx, dy) are computed and stored in a set to avoid duplicates.
4. **Chain Calculation**: For each (dx, dy) pair, the code checks each point to see if it can start a new chain by verifying if the previous point (x-dx, y-dy) exists in the set of points. The count of such starting points gives the number of chains.
5. **Find Minimum**: The minimum number of chains across all (dx, dy) pairs is determined and printed as the result.
\end{lstlisting}
\end{tcolorbox}

\begin{tcolorbox}[SMEPOfullexpert,title={Expert Output}]
\begin{lstlisting}[style=SMEPOfulltrace]
``\`python
n = int(input())
points = [tuple(map(int, input().split())) for _ in range(n)]

if n == 1:
    print(1)
    exit()

points_set = set(points)
diff_set = set()

for i in range(n):
    x_i, y_i = points[i]
    for j in range(n):
        if i != j:
            x_j, y_j = points[j]
            dx = x_j - x_i
            dy = y_j - y_i
            diff_set.add((dx, dy))

min_chains = None

for dx, dy in diff_set:
    current = 0
    for (x, y) in points:
        if (x - dx, y - dy) not in points_set:
            current += 1
    if min_chains is None or current < min_chains:
        min_chains = current

print(min_chains)
``\`
\end{lstlisting}
\end{tcolorbox}

\begin{tcolorbox}[SMEPOfullrl,title={GRPO Output}]
\begin{lstlisting}[style=SMEPOfulltrace]
``\` python
import sys

def main():
    N = int(input())
    balls = []
    for _ in range(N):
        balls.append(list(map(int, input().split())))
    
    min_cost = float('inf')
    for p in range(-10, 11):
        for q in range(-10, 11):
            if p == 0 and q == 0:
                continue
            cost = 0
            collected = set()
            for i in range(N):
                for j in range(i + 1, N):
                    if (balls[i][0], balls[i][1]) in collected and (balls[j][0], balls[j][1]) in collected:
                        continue
                    if (balls[i][0] - balls[j][0]) != p * (balls[i][1] - balls[j][1]) // q or (balls[i][1] - balls[j][1]) % q != 0:
                        cost += 1
                        collected.add((balls[i][0], balls[i][1]))
                        collected.add((balls[j][0], balls[j][1]))
            min_cost = min(min_cost, cost)
    print(min_cost)

if __name__ == "__main__":
    main()
``\`
\end{lstlisting}
\end{tcolorbox}

\begin{tcolorbox}[SMEPOfullSMEPO,title={SMEPO Output}]
\begin{lstlisting}[style=SMEPOfulltrace]
``\`python
import sys
import numpy as np

def main():
    n = int(sys.stdin.readline())
    points = [list(map(int, sys.stdin.readline().split())) for _ in range(n)]
    points_set = set(tuple(point) for point in points)

    if n == 1:
        print(1)
        return

    min_cost = float('inf')
    dx_dy_pairs = set()

    for i in range(n):
        for j in range(i+1, n):
            dx, dy = points[j][0] - points[i][0], points[j][1] - points[i][1]
            dx_dy_pairs.add((dx, dy))

    for dx, dy in dx_dy_pairs:
        chains = 0
        for x, y in points:
            if (x - dx, y - dy) not in points_set:
                chains += 1
        min_cost = min(min_cost, chains)

    print(min_cost)

if __name__ == "__main__":
    main()
``\`
\end{lstlisting}
\end{tcolorbox}

\begin{tcolorbox}[SMEPOfulloutcome,title={Outcome}]
\textbf{Expert:} Correct\quad
\textbf{GRPO:} Wrong\quad
\textbf{SMEPO:} Correct
\end{tcolorbox}

\subsubsection*{Full Rollout:  Entity-Specific Web Search}

\begin{tcolorbox}[SMEPOfullproblem,title={Question}]
\begin{lstlisting}[style=SMEPOfulltrace]
What age was the American social entrepeneur who was one of eight who got killed in 2015 when an Amtrak train travelling from Washington, D.C. to New York City derailed?
\end{lstlisting}
\end{tcolorbox}

\begin{tcolorbox}[SMEPOfullproblem,title={Ground Truth}]
\begin{lstlisting}[style=SMEPOfulltrace]
39
\end{lstlisting}
\end{tcolorbox}

\begin{tcolorbox}[SMEPOfullexpert,title={Unmasked Expert Hint}]
\begin{lstlisting}[style=SMEPOfulltrace]
Round 1: Let's break this down step by step:

1. Identify the Amtrak train derailment incident in 2015 involving a train from Washington, D.C. to New York City
2. Find the list of eight victims killed in that incident
3. Determine which of those victims was an American social entrepreneur
4. Find the age of that individual at the time of the incident

First, I'll search for information about the 2015 Amtrak derailment to identify the victims.

<search> Find information about the 2015 Amtrak train derailment from Washington, D.C. to New York City, including the list of victims killed in the incident. </search>
Round 2: From the list of victims, Rachel Jacobs was identified as the CEO of a technology education company, which aligns with the description of a social entrepreneur. Next, I'll search for her age at the time of the incident.

<search> Find the age of Rachel Jacobs at the time of the 2015 Amtrak train derailment. </search>
Round 3: Based on the information, Rachel Jacobs was 39 years old at the time of the 2015 Amtrak train derailment. She was the American social entrepreneur among the victims.

<answer>39</answer>
\end{lstlisting}
\end{tcolorbox}

\begin{tcolorbox}[SMEPOfullSMEPO,title={Masked Expert Hint}]
\begin{lstlisting}[style=SMEPOfulltrace]
Round 1: Let's break this down step by step:

1. Identify the [ENTITY] train derailment incident in [ENTITY] involving a train from [ENTITY], [ENTITY] to [ENTITY]
2. Find the list of [ENTITY] victims killed in that incident
3. Determine which of those victims was an [ENTITY] social entrepreneur
4. Find the age of that individual at the time of the incident

First, I'll search for information about the [ENTITY] [ENTITY] derailment to identify the victims.

<search> Find information about the [ENTITY] [ENTITY] train derailment from [ENTITY], [ENTITY] to [ENTITY], including the list of victims killed in the incident. </search>
Round 2: From the list of victims, [ENTITY] was identified as the CEO of a technology education company, which aligns with the description of a social entrepreneur. Next, I'll search for her age at the time of the incident.

<search> Find the age of [ENTITY] at the time of the [ENTITY] [ENTITY] train derailment. </search>
Round 3: Based on the information, [ENTITY] was [ENTITY] years old at the time of the [ENTITY] [ENTITY] train derailment. She was the [ENTITY] social entrepreneur among the victims.

<answer>[ENTITY]</answer>
\end{lstlisting}
\end{tcolorbox}

\begin{tcolorbox}[SMEPOfullexpert,title={Expert Output}]
\begin{lstlisting}[style=SMEPOfulltrace]
Answer: <answer>39</answer>
\end{lstlisting}
\end{tcolorbox}

\begin{tcolorbox}[SMEPOfullrl,title={GRPO Output}]
\begin{lstlisting}[style=SMEPOfulltrace]
To answer this question, I need to identify the American social entrepreneur who was killed in the Amtrak derailment in 2015 and then determine their age at the time of the accident.

First, I will search for information about the Amtrak derailment in 2015 to find out who was involved.

<search> Amtrak derailment 2015 </search>

<information>Doc 1(Title: "2015 Philadelphia train derailment") 2015 Philadelphia train derailment On May 12, 2015, an Amtrak ""Northeast Regional"" train from Washington, D.C. bound for New York City derailed and wrecked on the Northeast Corridor in the Port Richmond neighborhood of Philadelphia, Pennsylvania. Of 238 passengers and 5 crew on board, 8 were killed and over 200 injured, 11 critically. The train was traveling at in a zone of curved tracks when it derailed. Some of the passengers had to be extricated from the wrecked cars. Many of the passengers and local residents helped first responders during the rescue operation. Five local hospitals treated the injured. The
Doc 2(Title: "2015 Philadelphia train derailment") 6, 2018, following an appeal of Gehret's ruling by the Pennsylvania Attorney General's Office, Philadelphia Common Pleas Court Judge Kathryn Streeter Lewis reinstated all criminal charges against Bostian. 2015 Philadelphia train derailment On May 12, 2015, an Amtrak ""Northeast Regional"" train from Washington, D.C. bound for New York City derailed and wrecked on the Northeast Corridor in the Port Richmond neighborhood of Philadelphia, Pennsylvania. Of 238 passengers and 5 crew on board, 8 were killed and over 200 injured, 11 critically. The train was traveling at in a zone of curved tracks when it derailed. Some of the passengers had
Doc 3(Title: "2015 Philadelphia train derailment") the curve at which the derailment occurred, and that also would have limited the train's speed entering the curve. Shortly after the derailment, Amtrak completed ATC installation on the northbound track. The 2015 wreck was the deadliest on the Northeast Corridor since 1987, when 16 people died in a wreck near Baltimore. At about 9:10 p.m. (EDT) on Tuesday, May 12, 2015, Amtrak's northbound ""Northeast Regional"" 188 departed Philadelphia's 30th Street Station en route from Washington, D.C., to New York City. The train consisted of seven cars hauled by a year-old Amtrak Cities Sprinter (ACS)-64 locomotive, No. 601. The engineer</information>

From Doc 1, I see that the Amtrak derailment in 2015 involved a train traveling from Washington, D.C. to New York City and resulted in 8 deaths and over 200 injuries. However, I don't see the names of the individuals who were killed. I will need to search further to find the specific individuals involved in the accident.

<search> Amtrak derailment 2015 victims </search>

<information>Doc 1(Title: "2015 Philadelphia train derailment") 2015 Philadelphia train derailment On May 12, 2015, an Amtrak ""Northeast Regional"" train from Washington, D.C. bound for New York City derailed and wrecked on the Northeast Corridor in the Port Richmond neighborhood of Philadelphia, Pennsylvania. Of 238 passengers and 5 crew on board, 8 were killed and over 200 injured, 11 critically. The train was traveling at in a zone of curved tracks when it derailed. Some of the passengers had to be extricated from the wrecked cars. Many of the passengers and local residents helped first responders during the rescue operation. Five local hospitals treated the injured. The
Doc 2(Title: "2015 Philadelphia train derailment") 6, 2018, following an appeal of Gehret's ruling by the Pennsylvania Attorney General's Office, Philadelphia Common Pleas Court Judge Kathryn Streeter Lewis reinstated all criminal charges against Bostian. 2015 Philadelphia train derailment On May 12, 2015, an Amtrak ""Northeast Regional"" train from Washington, D.C. bound for New York City derailed and wrecked on the Northeast Corridor in the Port Richmond neighborhood of Philadelphia, Pennsylvania. Of 238 passengers and 5 crew on board, 8 were killed and over 200 injured, 11 critically. The train was traveling at in a zone of curved tracks when it derailed. Some of the passengers had
Doc 3(Title: "2015 Philadelphia train derailment") the curve at which the derailment occurred, and that also would have limited the train's speed entering the curve. Shortly after the derailment, Amtrak completed ATC installation on the northbound track. The 2015 wreck was the deadliest on the Northeast Corridor since 1987, when 16 people died in a wreck near Baltimore. At about 9:10 p.m. (EDT) on Tuesday, May 12, 2015, Amtrak's northbound ""Northeast Regional"" 188 departed Philadelphia's 30th Street Station en route from Washington, D.C., to New York City. The train consisted of seven cars hauled by a year-old Amtrak Cities Sprinter (ACS)-64 locomotive, No. 601. The engineer</information>

From Doc 1, I see that 8 people were killed in the derailment, but I still don't have the names of the individuals. I will need to search for a more detailed list of the victims.

<search> names of the 8 people killed in the 2015 Amtrak derailment </search>

<information>Doc 1(Title: "2015 Philadelphia train derailment") 2015 Philadelphia train derailment On May 12, 2015, an Amtrak ""Northeast Regional"" train from Washington, D.C. bound for New York City derailed and wrecked on the Northeast Corridor in the Port Richmond neighborhood of Philadelphia, Pennsylvania. Of 238 passengers and 5 crew on board, 8 were killed and over 200 injured, 11 critically. The train was traveling at in a zone of curved tracks when it derailed. Some of the passengers had to be extricated from the wrecked cars. Many of the passengers and local residents helped first responders during the rescue operation. Five local hospitals treated the injured. The
Doc 2(Title: "2015 Philadelphia train derailment") the curve at which the derailment occurred, and that also would have limited the train's speed entering the curve. Shortly after the derailment, Amtrak completed ATC installation on the northbound track. The 2015 wreck was the deadliest on the Northeast Corridor since 1987, when 16 people died in a wreck near Baltimore. At about 9:10 p.m. (EDT) on Tuesday, May 12, 2015, Amtrak's northbound ""Northeast Regional"" 188 departed Philadelphia's 30th Street Station en route from Washington, D.C., to New York City. The train consisted of seven cars hauled by a year-old Amtrak Cities Sprinter (ACS)-64 locomotive, No. 601. The engineer
Doc 3(Title: "2015 Philadelphia train derailment") 6, 2018, following an appeal of Gehret's ruling by the Pennsylvania Attorney General's Office, Philadelphia Common Pleas Court Judge Kathryn Streeter Lewis reinstated all criminal charges against Bostian. 2015 Philadelphia train derailment On May 12, 2015, an Amtrak ""Northeast Regional"" train from Washington, D.C. bound for New York City derailed and wrecked on the Northeast Corridor in the Port Richmond neighborhood of Philadelphia, Pennsylvania. Of 238 passengers and 5 crew on board, 8 were killed and over 200 injured, 11 critically. The train was traveling at in a zone of curved tracks when it derailed. Some of the passengers had</information>

From Doc 1, I still don't have the names of the individuals. I will need to search for a more detailed list of the victims.

<search> 2015 Amtrak derailment victim list </search>

<information>Doc 1(Title: "2015 Philadelphia train derailment") 2015 Philadelphia train derailment On May 12, 2015, an Amtrak ""Northeast Regional"" train from Washington, D.C. bound for New York City derailed and wrecked on the Northeast Corridor in the Port Richmond neighborhood of Philadelphia, Pennsylvania. Of 238 passengers and 5 crew on board, 8 were killed and over 200 injured, 11 critically. The train was traveling at in a zone of curved tracks when it derailed. Some of the passengers had to be extricated from the wrecked cars. Many of the passengers and local residents helped first responders during the rescue operation. Five local hospitals treated the injured. The
Doc 2(Title: "2015 Philadelphia train derailment") 6, 2018, following an appeal of Gehret's ruling by the Pennsylvania Attorney General's Office, Philadelphia Common Pleas Court Judge Kathryn Streeter Lewis reinstated all criminal charges against Bostian. 2015 Philadelphia train derailment On May 12, 2015, an Amtrak ""Northeast Regional"" train from Washington, D.C. bound for New York City derailed and wrecked on the Northeast Corridor in the Port Richmond neighborhood of Philadelphia, Pennsylvania. Of 238 passengers and 5 crew on board, 8 were killed and over 200 injured, 11 critically. The train was traveling at in a zone of curved tracks when it derailed. Some of the passengers had
Doc 3(Title: "2015 Philadelphia train derailment") the curve at which the derailment occurred, and that also would have limited the train's speed entering the curve. Shortly after the derailment, Amtrak completed ATC installation on the northbound track. The 2015 wreck was the deadliest on the Northeast Corridor since 1987, when 16 people died in a wreck near Baltimore. At about 9:10 p.m. (EDT) on Tuesday, May 12, 2015, Amtrak's northbound ""Northeast Regional"" 188 departed Philadelphia's 30th Street Station en route from Washington, D.C., to New York City. The train consisted of seven cars hauled by a year-old Amtrak Cities Sprinter (ACS)-64 locomotive, No. 601. The engineer</information>

From Doc 1, I still don't have the names of the individuals. I will need to search for a more detailed list of the victims.

<search> 2015 Philadelphia train derailment victim list </search>

<information>Doc 1(Title: "2015 Philadelphia train derailment") 2015 Philadelphia train derailment On May 12, 2015, an Amtrak ""Northeast Regional"" train from Washington, D.C. bound for New York City derailed and wrecked on the Northeast Corridor in the Port Richmond neighborhood of Philadelphia, Pennsylvania. Of 238 passengers and 5 crew on board, 8 were killed and over 200 injured, 11 critically. The train was traveling at in a zone of curved tracks when it derailed. Some of the passengers had to be extricated from the wrecked cars. Many of the passengers and local residents helped first responders during the rescue operation. Five local hospitals treated the injured. The
Doc 2(Title: "2015 Philadelphia train derailment") 6, 2018, following an appeal of Gehret's ruling by the Pennsylvania Attorney General's Office, Philadelphia Common Pleas Court Judge Kathryn Streeter Lewis reinstated all criminal charges against Bostian. 2015 Philadelphia train derailment On May 12, 2015, an Amtrak ""Northeast Regional"" train from Washington, D.C. bound for New York City derailed and wrecked on the Northeast Corridor in the Port Richmond neighborhood of Philadelphia, Pennsylvania. Of 238 passengers and 5 crew on board, 8 were killed and over 200 injured, 11 critically. The train was traveling at in a zone of curved tracks when it derailed. Some of the passengers had
Doc 3(Title: "2015 Philadelphia train derailment") by the derailment in 2015. This amount is in response to over 125 cases filed against Amtrak by passengers and family members, but can also be made available to passengers who had not yet filed a lawsuit or settled with Amtrak. On May 9, 2017, Philadelphia prosecutors said Bostian, the train's engineer, would not face any criminal charges because there was no evidence that he had acted with criminal intent. On May 12, 2017, on the anniversary of the crash, Philadelphia Municipal Court Judge Marsha Neifield ordered prosecutors to file involuntary manslaughter and reckless endangerment charges against Bostian; the district</information>

From Doc 1, I still don't have the names of the individuals. I will need to search for a more detailed list of the victims.

<search> 2015 Philadelphia train derailment victim names </search>
\end{lstlisting}
\end{tcolorbox}

\begin{tcolorbox}[SMEPOfullSMEPO,title={SMEPO Output}]
\begin{lstlisting}[style=SMEPOfulltrace]
To answer this question, I need to identify the American social entrepreneur who was one of the eight killed in the 2015 Amtrak train derailment from Washington, D.C. to New York City, and determine her age at the time of the incident.

First, I'll search for information about the Amtrak train derailment in 2015 involving a train from Washington, D.C. to New York City, including the list of victims killed in the incident.

<search> Find information about the 2015 Amtrak train derailment from Washington, D.C. to New York City, including the list of victims killed in the incident. </search>

<information>Doc 1(Title: "2015 Philadelphia train derailment") 2015 Philadelphia train derailment On May 12, 2015, an Amtrak ""Northeast Regional"" train from Washington, D.C. bound for New York City derailed and wrecked on the Northeast Corridor in the Port Richmond neighborhood of Philadelphia, Pennsylvania. Of 238 passengers and 5 crew on board, 8 were killed and over 200 injured, 11 critically. The train was traveling at in a zone of curved tracks when it derailed. Some of the passengers had to be extricated from the wrecked cars. Many of the passengers and local residents helped first responders during the rescue operation. Five local hospitals treated the injured. The
Doc 2(Title: "2015 Philadelphia train derailment") 6, 2018, following an appeal of Gehret's ruling by the Pennsylvania Attorney General's Office, Philadelphia Common Pleas Court Judge Kathryn Streeter Lewis reinstated all criminal charges against Bostian. 2015 Philadelphia train derailment On May 12, 2015, an Amtrak ""Northeast Regional"" train from Washington, D.C. bound for New York City derailed and wrecked on the Northeast Corridor in the Port Richmond neighborhood of Philadelphia, Pennsylvania. Of 238 passengers and 5 crew on board, 8 were killed and over 200 injured, 11 critically. The train was traveling at in a zone of curved tracks when it derailed. Some of the passengers had
Doc 3(Title: "2015 Philadelphia train derailment") the curve at which the derailment occurred, and that also would have limited the train's speed entering the curve. Shortly after the derailment, Amtrak completed ATC installation on the northbound track. The 2015 wreck was the deadliest on the Northeast Corridor since 1987, when 16 people died in a wreck near Baltimore. At about 9:10 p.m. (EDT) on Tuesday, May 12, 2015, Amtrak's northbound ""Northeast Regional"" 188 departed Philadelphia's 30th Street Station en route from Washington, D.C., to New York City. The train consisted of seven cars hauled by a year-old Amtrak Cities Sprinter (ACS)-64 locomotive, No. 601. The engineer</information>

From the information, I've found that the Amtrak train derailment occurred on May 12, 2015, and 8 people were killed, including a social entrepreneur. Next, I'll search for the name of the social entrepreneur and her age at the time of the incident.

<search> Find the name of the American social entrepreneur who was one of the eight killed in the 2015 Amtrak train derailment from Washington, D.C. to New York City, and her age at the time of the incident. </search>

<information>Doc 1(Title: "Rachel Jacobs") Rachel Jacobs Rachel Jacobs (October 3, 1975 -- May 12, 2015) was an American social entrepreneur and CEO of a tech company. She was killed at age 39 in the 2015 Philadelphia train derailment while commuting between her home in New York and the Philadelphia offices of ApprenNet, the educational technology company she had recently joined as CEO. Jacobs grew up in Huntington Woods, Michigan, the daughter of Gilda Jacobs, a former Michigan state senator. She was a 1993 graduate of Berkley High School, a 1997 graduate of Swarthmore College, and a 2002 graduate of Columbia Business School. Jacobs moved
Doc 2(Title: "Rachel Jacobs") Safety Board to prevent future tragedies in which a two-year-old child ""grows up without a mother."" Assigned to ride on the first train through Philadelphia after the derailment, columnist Ronnie Polaneczky wrote for ""The Philadelphia Inquirer"" that, ""The next time I sigh that I can't afford to fix the roof or haven't time to help a friend move across the country, I will try to remind myself that Rachel Jacobs would have given anything to still be here to indulge such petty worries."" Rachel Jacobs Rachel Jacobs (October 3, 1975 -- May 12, 2015) was an American social entrepreneur and
Doc 3(Title: "Melissa Helmbrecht") several awards, including the White House Building Healthy Communities and Healthy Youth Award, the CBS Everyday Hero Award, the Walt Disney World Dreamers and Doers Award, and the National Caring Award. She was also inducted into the Frederick Douglass Museum and Hall of Fame for Caring Americans and was named ""One of the Six Leading Social Entrepreneurs in America"" by Youth Service America. Melissa Helmbrecht Melissa Helmbrecht ( 1975) is a New Jersey based social entrepreneur and activist who over the course of her career works to solve social problems through nonprofit work. Over the course of her career she</information>

Based on the information, the American social entrepreneur who was one of the eight killed in the 2015 Amtrak train derailment from Washington, D.C. to New York City was Rachel Jacobs. She was 39 years old at the time of the incident.

<answer>39</answer>
\end{lstlisting}
\end{tcolorbox}

\begin{tcolorbox}[SMEPOfulloutcome,title={Outcome}]
\textbf{Expert:} Correct\quad
\textbf{GRPO:} Wrong\quad
\textbf{SMEPO:} Correct
\end{tcolorbox}

\section{Additional Training and Evaluation Details}
\label{app:training-eval-details}
\subsection{Training Details}
\label{app:training-details}

\paragraph{Training data.}
We conduct training across three domains: mathematical reasoning, code generation, and agentic search. For mathematical reasoning and code generation, we construct fixed 1k-example hard subsets from OpenThoughts-derived math and code training pools \citep{openthoughts}. To construct these subsets, we run Qwen3-4B-Instruct with four rollouts per problem and select examples for which all four rollouts are incorrect. This filtering is used only to obtain difficult training instances with reliable room for improvement; after selection, the same subset is shared by all compared methods. Each retained example contains the original task prompt, the reference answer or verifier, and an optional expert trace.

For agentic search, we construct a fixed 1k-example subset from MiroVerse-HotpotQA \citep{miroverse}. We first filter for examples whose final answer can be reliably extracted, and then randomly sample 1k examples from the filtered pool. This filtering ensures that the reward function can be applied consistently. Each retained example contains the user question, the search-interaction format, the reference answer, and an optional expert search trajectory.

In each domain, full-expert-trace, prefix-trace, random-mask, and \method{} variants are constructed from the same underlying examples and expert traces. Thus, compared variants differ only in the guidance field, while the original task prompt, reference answer, verifier, and number of training examples are held fixed.

\paragraph{Expert-trace preprocessing.}
\label{app:masking-details}
Expert traces are preprocessed offline before training. For RL-based guided training, each example is converted into several guidance variants constructed from the same original expert trace. The full-expert-trace variant exposes the complete expert trace. The prefix-trace variant exposes only an initial prefix of the trace. The random-mask variant replaces randomly selected spans with mask placeholders while matching the masking budget of \method{} whenever applicable. \method{} applies domain-specific masking rules that remove reward-relevant content while preserving the surrounding reasoning, program, or interaction semantics.

For math, we mask numerical content in the expert trace. Specifically, we replace standalone numeric spans, including decimal numbers and other typical numeric spans, with a special \texttt{[NUMBER]} placeholder before appending the processed trace to the original problem prompt. The regular expression is designed not to mask digits that are part of step or case labels, such as \texttt{Step 1} and \texttt{Case 1}. This removes many intermediate and final numerical values that can directly reveal the answer, while preserving surrounding algebraic transformations, natural-language reasoning, and non-answer-relevant labels.

For code, we mask code blocks in the expert trace and replace the entire code body with a single \texttt{[CODE]} placeholder. For example, a Python fenced block keeps the opening fence and language tag, replaces the body with \texttt{[CODE]}, and keeps the closing fence. We apply this preprocessing to all available expert-trace fields. This removes directly executable expert solutions while retaining the surrounding natural-language reasoning and the information that a code solution is required.

For agentic search, we mask named-entities over the expert search trace. Concretely, we use spaCy~\citep{spacy}, a standard named-entity recognition tool, to detect named entities such as people, organizations, and locations in the trace. In addition, we extract answer spans and terms explicitly emphasized with bold markup, since these spans often correspond to final-answer entities, intermediate entities, query pivots, or supporting-page titles. We then replace all exact same-name occurrences of the detected or extracted entity strings within the entire trace, including the first occurrence, with \texttt{[ENTITY]}. The masked trace is appended to the original prompt, preserving the multi-step search trajectory while hiding critical entity cues that would otherwise reveal the search path or final answer.

For SFT, we use complete expert traces as supervision targets. For SFT+RL experiments in math tasks, the SFT stage also follows this complete-trace supervision setup. All preprocessed guidance inputs are generated once and kept fixed throughout training.

\paragraph{Implementation framework.}
We implement the math and code SFT/RL pipelines in VeRL \citep{verl}. For SFT, we use VeRL's FSDP-based supervised fine-tuning trainer. For RLVR training, we use VeRL's standard components for rollout generation, reward computation, advantage estimation, and policy optimization, with vLLM as the rollout backend. For agentic search, we use the Search-R1 training framework \cite{searchr1}, which is also built on a VeRL-style RL pipeline but includes its own search-interaction environment, trajectory generation logic, and task-specific reward interface. The SFT stage for agentic search is implemented with the same VeRL-based SFT pipeline as the other domains.

Importantly, within each domain, \method{} does not modify the underlying SFT loss, RL optimizer, reward function, verifier, rollout backend, or rollout engine. All compared variants in the same domain therefore share the same training code path for the corresponding stage, distributed execution backend, reward computation, and evaluation interface. The only difference across full-expert-trace, prefix-trace, random-mask, and \method{} variants is the form of expert guidance exposed to the policy. This design ensures that the observed differences are attributable to the guidance format rather than to changes in the training framework or optimization procedure.

\paragraph{Training objective and reward.}
All RL runs use a GRPO-style RLVR objective. Expert guidance is included only as part of the policy input and is not used directly in reward computation. For math and code, rewards are computed by task-specific verifiers through VeRL's data-source-specific \texttt{default\_compute\_score} interface. For math datasets, the corresponding data sources are dispatched to \texttt{math\_reward.compute\_score}, which extracts the model's final answer, applies answer normalization, and compares it with the reference answer \citep{verl}. For code datasets, the corresponding data sources are dispatched to a PrimeCode-style non-continuous binary verifier: a generated program receives reward 1 only if it passes all associated test cases, and 0 otherwise \citep{verl}.

For agentic-search, we use the reward implementation from the Search-R1 training codebase, which is dispatched to \texttt{qa\_em.compute\_score\_em}. This function extracts the final answer from the model output using the \texttt{<answer>...</answer>} format, normalizes both the prediction and the reference answer by lowercasing, removing punctuation and articles, and fixing whitespace, and then assigns reward 1 for exact match and 0 otherwise \citep{searchr1}.

\paragraph{Hyperparameters and compute.}
We use three different configurations: one shared SFT configuration, one RL configuration for math and code, and one RL configuration for agentic search. Within each domain, base model, and training stage, all compared variants use the same data split, compute setup, prompt template, optimizer settings, rollout configuration when applicable, and reward function. 

For SFT, all domains use the same VeRL-based supervised fine-tuning setup. Training is conducted with bfloat16 precision and FSDP on a single node with 8 H100 GPUs unless otherwise stated. Pure SFT runs do not require rollout generation.

For math and code RL, we use VeRL's RLVR pipeline with vLLM as the rollout backend. Training is conducted with bfloat16 precision and FSDP on a single node with 8 H100 GPUs. For RL-based comparisons, the reward function, rollout configuration, GRPO hyperparameters, and verifier are fixed across variants.

For agentic-search RL, we use the Search-R1 training setup with a GRPO-style objective and vLLM as the rollout backend. Training is conducted on a single node with 8 H100 GPUs. Compared with math and code RL, agentic-search RL additionally includes an interactive search environment, multi-turn trajectory generation, and a local retriever. Within agentic search, the compute setup, search backend, rollout configuration, reward function, GRPO hyperparameters, and data split are fixed across variants.

For pure RL and pure SFT runs, we train on the full training set for the corresponding domain. For SFT+RL runs in math tasks, we split the same training set into two disjoint halves: the first half is used for SFT, and the second half is used for the subsequent RL stage. This ensures that SFT+RL does not use more training examples than the corresponding single-stage baselines. The RL stage in SFT+RL is initialized from the resulting SFT checkpoint and uses the same RL configuration as the corresponding pure RL run. Table~\ref{tab:shared-compute-setup} summarizes the shared compute and data setup, while Table~\ref{tab:config} reports the hyperparameters.

\begin{table*}[t]
\centering
\small
\setlength{\tabcolsep}{4.5pt}
\renewcommand{\arraystretch}{1.05}
\begin{tabular}{lccc}
\toprule
Setting & Math & Code & Search \\
\midrule
Framework & VeRL & VeRL & Search-R1 \\
Hardware & 8 H100 & 8 H100 & 8 H100 \\
Precision & bf16 & bf16 & bf16 \\
Update backend & FSDP & FSDP & FSDP \\
Base model & Qwen3-8B-Base & Qwen3-8B-Base & Qwen3-8B-Base \\
Train examples & 1,024 & 1,024 & 1,024 \\
Training source & OpenThoughts & OpenThoughts & MiroVerse-HotpotQA \\
Expert source & DeepSeek-R1 & DeepSeek-R1 & MiroVerse-HotpotQA trajectories \\
\bottomrule
\end{tabular}
\caption{\textbf{Shared compute and data setup for the main training experiments.} Math and code examples are selected from OpenThoughts-derived training pools, with DeepSeek-R1 expert traces. Agentic-search examples and expert trajectories are selected from MiroVerse-HotpotQA.}
\label{tab:shared-compute-setup}
\end{table*}
\begin{table*}[t]
\centering
\small
\setlength{\tabcolsep}{4.2pt}
\renewcommand{\arraystretch}{1.04}
\begin{tabular}{lccc}
\toprule
Hyperparameter & SFT & Math/Code RL & Search RL \\
\midrule
Framework & VeRL & VeRL & Search-R1 \\
Backend & N/A & vLLM & vLLM \\
Reward/verifier & N/A & Task-specific & Exact-match \\
Remove padding & Enabled & Enabled & Disabled \\
Learning rate & $1\times10^{-5}$ & $1\times10^{-6}$ & $8\times10^{-7}$ \\
Warmup ratio & 0.1 & 0.1 & 0.285 \\
Weight decay & 0.01 & 0.1 & $1\times10^{-2}$ \\
Adam $\beta_1,\beta_2$ & 0.9, 0.95 & 0.9, 0.999 & 0.9, 0.999 \\
Gradient clipping & 1.0 & 1.0 & 1.0 \\
LR scheduler & Cosine & Cosine & Constant \\
Training steps & 1000 & 400 & 400 \\
Train batch size & 256 & 32 & 32 \\
Rollouts per prompt & N/A & 8 & 8 \\
Max sequence length & 16,384 & N/A & N/A \\
Max prompt / response length & N/A & 4,096 / 12,288 & 4,096 / 2,048 \\
Max start / observation length & N/A & N/A & 2,048 / 2,048 \\
Max search turns & N/A & N/A & 5 \\
Retriever top-$k$ & N/A & N/A & 3 \\
Advantage estimator & N/A & GRPO & GRPO \\
KL coefficient & N/A & 0.0 & 0.02 \\
KL loss type & N/A & N/A & Low-var KL \\
Policy clip ratio & N/A & 0.0, 0.5 & 0.2 \\
Entropy coefficient & N/A & 0.0 & 0.001 \\
Rollout temperature & N/A & 0.8 & 0.8 \\
Rollout top-$p$ & N/A & 1.0 & 0.95 \\
\bottomrule
\end{tabular}
\caption{\textbf{Hyperparameters for SFT, math/code RL, and agentic-search RL.} Pure SFT uses only the SFT configuration. Math and code RL use the same VeRL RLVR configuration. Search RL uses the Search-R1 setup with vLLM rollout, multi-turn search interaction, and a local retriever. SFT+RL first performs SFT on half of the training set and then runs RL on the other half, initialized from the SFT checkpoint.}
\label{tab:config}
\end{table*}

\subsection{Evaluation Details}
\label{app:eval-details}

We evaluate each benchmark with its corresponding verifier or official evaluation pipeline, and additionally report MMLU as a general-capability diagnostic.

\paragraph{Math evaluation.}
For mathematical reasoning, we evaluate on six benchmarks: GSM8K \citep{gsm8k}, MATH500 \citep{math500}, AIME25 \citep{aime25}, AIME26 \citep{aime26}, AMC \citep{amc23}, and OlympiadBench \citep{olympiadbench}. We use a verifier adapted from lm-evaluation-harness \citep{eval-harness}: the evaluator extracts the model's final answer, applies answer normalization, and compares it with the reference answer. For AIME25 and AIME26, each problem is repeated 32 times in the evaluation set; we generate one response for each repeated instance, score each response independently, and report the overall accuracy over all repeated instances. We report accuracy on each benchmark and the unweighted average across the six benchmarks.

\paragraph{Code evaluation.}
For code generation, we evaluate on HumanEval and HumanEval+ using lm-evaluation-harness with the vLLM backend \citep{eval-harness}. We additionally evaluate on LiveCodeBench with the official LiveCodeBench runner \citep{lcb} under the code-generation scenario. Both evaluation pipelines use execution-based pass@1 scoring: a generated program is counted as correct only if it passes the corresponding test cases. For the main results in Section~\ref{sec:experiments}, we use the \texttt{release\_v1} split of LiveCodeBench.

\paragraph{Agentic-search evaluation.}
For agentic-search tasks, we evaluate on Bamboogle \citep{bamboogle} using the Search-R1 evaluation pipeline, which uses a normalized exact-match metric. The model generates its own multi-turn search trajectory using the same search-interaction format as training. The final answer is extracted from the model output and scored by normalized exact match against the reference answer, using the same answer-normalization convention as the Search-R1 reward function.

\paragraph{General-capability evaluation.}
To check whether domain-specific training harms general ability, we additionally evaluate all models on MMLU \citep{mmlu_1,mmlu_2} using lm-evaluation-harness \citep{eval-harness}. We use the 5-shot log-likelihood setting and report the aggregate MMLU score. This evaluation is separate from the task-specific verifier evaluations above and serves as a general-capability diagnostic.

\section{Limitations and Future Work}
\label{app:limitations}

SMEPO is designed as a general masked-guidance principle for RLVR settings where expert traces may expose outcome-informative content, including final answers and intermediate spans along the critical path to obtaining them. Due to computational and engineering constraints, our empirical study focuses on three representative domains: mathematical reasoning, code generation, and search-based agents. These domains cover different forms of outcome-informative content, including numerical values, executable programs, and answer-relevant entities, but they do not exhaust the full space of RLVR tasks and the corresponding verifier designs.

A natural direction for future work is to evaluate SMEPO on broader and evolving RLVR benchmarks, and to develop more automatic ways to identify outcome-informative spans under new task formats and verifier semantics. This may involve combining lightweight heuristics with stronger semantic parsers or learned span detectors, especially as RLVR evaluation increasingly includes more complex agentic tasks.

\section{Broader Impacts}
\label{app:broader-impacts}

This work studies a training method for expert-guided RLVR. Its main potential benefit is to improve the effectiveness of expert-guided RL training by reducing trace-induced reward hacking: SMEPO masks reward-informative spans so that the policy is less likely to exploit exposed answers, executable programs, or relevant entities.

The potential negative impacts are indirect. More capable reasoning and code-generation models could be misused in similar ways to other generative models, such as producing incorrect but persuasive outputs or assisting unsafe code generation. However, this paper does not release a deployed system, a new high-risk model, or a dataset designed for harmful use; our experiments are conducted on standard academic benchmarks. Models trained with SMEPO should still undergo standard safety and robustness evaluations before deployment, especially in settings involving code execution, tool use, or external information access.

\end{document}